# From Text to Insight: Leveraging Large Language Models for Performance Evaluation in Management

Ning Li
Tsinghua University
lining@sem.tsinghua.edu.cn

Huaikang Zhou
Tsinghua University
zhouhk@sem.tsinghua.edu.cn

Mingze Xu
Tsinghua University
xumz22@mails.tsinghua.edu.cn

**Abstract**

This study explores the potential of Large Language Models (LLMs), specifically GPT-4, to enhance objectivity in organizational task performance evaluations. Through comparative analyses across two studies, including various task performance outputs, we demonstrate that LLMs can serve as a reliable and even superior alternative to human raters in evaluating knowledge-based performance outputs, which are a key contribution of knowledge workers. Our results suggest that GPT ratings are comparable to human ratings but exhibit higher consistency and reliability. Additionally, combined multiple GPT ratings on the same performance output show strong correlations with aggregated human performance ratings, akin to the consensus principle observed in performance evaluation literature. However, we also find that LLMs are prone to contextual biases, such as the halo effect, mirroring human evaluative biases. Our research suggests that while LLMs are capable of extracting meaningful constructs from text-based data, their scope is currently limited to specific forms of performance evaluation. By highlighting both the potential and limitations of LLMs, our study contributes to the discourse on AI's role in management studies and sets a foundation for future research to refine AI's theoretical and practical applications in management.

In organizational research, understanding the variances in employee performance stands as a cornerstone, encompassing a broad spectrum from productivity and creativity to prosocial and deviant behaviors (Berry, Ones, & Sackett, 2007; Campbell & Wiernik, 2015; Organ & Ryan, 1995; Schleicher, Baumann, Sullivan, & Yim, 2019; Viswesvaran & Ones, 2000; Zhou & George, 2001). Traditionally, this endeavor has largely depended on evaluations conducted by human observers—leaders and coworkers—who assess the performance of focal employees (Borman, 1991). While necessary, this approach is not without its challenges. Primarily, it suffers from subjectivity and bias, relying on recollections of past behaviors over extended periods (Landy & Farr, 1980; Murphy & Cleveland, 1995). Moreover, evaluations often represent aggregated perceptions across diverse situations, mostly derived from a single source of rating, typically a direct supervisor (Woehr & Huffcutt, 1994). Such methodologies are inherently limited in their ability to capture the nuanced contributions of knowledge workers accurately (Scullen, Mount, & Goff, 2000; Viswesvaran & Ones, 2000).

Recognizing these limitations, there's a growing acknowledgment of the value inherent in knowledge-based performance outputs within management research (Spector & Pindek, 2016). These outputs, which timely capture the contributions of knowledge employees, offer a window into the granular aspects of performance (Berg, 2016, 2019; Perry-Smith & Mannucci, 2017; Sijbom, Janssen, & Van Yperen, 2015). For instance, within the domain of individual creativity, there has been a shift away from evaluating aggregated creative behaviors towards a finer grained analysis that zooms in on concrete, specific creative outputs—such as detailed plans, innovative ideas, and prototypes (Berg, 2016; Lu, Bartol, Venkataramani, Zheng, & Liu, 2019). This nuanced approach of evaluating creativity at the idea level has not only moved the field forward but also exemplified the broader utility of knowledge-based data in management research. By adopting this detailed analytical lens, we gain a more refined and precise understanding of key management constructs, showcasing the extensive applicability of text-based information to capture the meaningful constructs in organizational research beyond creativity (Carton, Murphy, & Clark, 2014; Maynes & Podsakoff, 2014; Sonenshein, 2010).

However, the logistical challenges of processing and extracting theoretically relevant constructs from the voluminous text-based outputs generated by knowledge workers have traditionally been daunting (Short, McKenny, & Reid, 2018). The labor-intensive, time-consuming, and costly nature of this process has often limited the use of human raters to laboratory settings or small-scale investigations. While previous Natural Language Processing (NLP) techniques show some promise in extracting text information for applications like sentiment analysis, they frequently struggle with the complex nature of management constructs such as performance, novelty, and creativity (George, Haas, & Pentland, 2014; Hannigan et al., 2019; Kobayashi, Mol, Berkers, Kismihók, & Den Hartog, 2017). Not only do these techniques necessitate extensive human involvement for pre-labeling and training, but they are also inherently restricted to a narrow scope of concepts (Harrison, Thurgood, Boivie, & Pfarrer, 2019; Tonidandel, King, & Cortina, 2018).

The development of LLMs like GPT-4 introduces new capabilities in processing nuanced text-based data, potentially overcoming the limitations faced by traditional methodologies (Brown et al., 2020; Vaswani et al., 2017). As a result, those technologies represent a potential transformative shift in how we can approach the extraction of meaningful constructs from textual data for organizational research. With their zero-shot learning capabilities—where models can make predictions on tasks they were not explicitly trained on—these models negate the need for pre-labeled data or extensive training. They require only clear construct definitions and prompts to extract pivotal management concepts—the "what" dimension fundamental to theoretical development (Ouyang et al., 2022; Whetten, 1989). Prompts act as instructions to LLMs, allowing them to process relevant information similarly to how humans perform tasks with instructions. This promising frontier, however, raises pivotal questions about the feasibility and efficacy of integrating LLMs into the organizational research toolkit (Shrestha, Ben-Menahem, & von Krogh, 2019).

First, how do the evaluations conducted by LLMs compare with those made by human judges in terms of accuracy and consistency (Brynjolfsson & Mitchell, 2017)? Second, do LLMs replicate known human biases in performance evaluations, such as the halo effect, or do they offer a more objective lens through which we can view employee contributions (Caliskan, Bryson, & Narayanan, 2017; Thorndike, 1920)? To address these questions, we conduct a comprehensive comparison, analyzing two sets of human-generated text-based performance outputs using both LLM and human evaluations. This design not only allows us to examine the relative merits of LLMs versus traditional human assessments but also to examine the nuances of how AI interprets and rates complex work-related constructs.

Our investigation shows compelling evidence of the feasibility and reliability of LLMs like GPT in assessing performance outputs. The analysis indicates a high degree of convergence between GPT ratings and the aggregated judgments of six human raters, which we use as a proxy for ground truth (Amabile, 1982; Kaufman, Baer, Cole, & Sexton, 2008). Across two studies, we evaluated a total of 520 task outputs in Study 1 and 224 task outputs in Study 2. Each task was assessed by a diverse and competent group of six judges, resulting in a total of 486 raters in Study 1 and 420 raters in Study 2. Notably, GPT demonstrates a remarkable consistency in providing reliable ratings across a spectrum of tasks, even those with varying levels of subjectivity and complexity. This consistency highlights GPT's potential to provide more accurate evaluations than individual human raters, demonstrating its viability as an alternative for performance assessment in organizational research.

Our investigation shows compelling evidence of the feasibility and reliability of LLMs like GPT in assessing performance outputs. The analysis indicates a high degree of convergence between GPT ratings and the aggregated judgments of six human raters, which we use as a proxy for ground truth (Amabile, 1982; Kaufman, Baer, Cole, & Sexton, 2008). Notably, GPT demonstrates a remarkable consistency in providing reliable ratings across a spectrum of tasks, even those with varying levels of subjectivity and complexity. This consistency highlights GPT's potential to provide more accurate evaluations than individual human raters, demonstrating its viability as an alternative for performance assessment in organizational research.

Moreover, our findings highlight the precision of GPT evaluations, evidenced by significantly less variance in ratings compared to human assessments. This precision suggests that LLMs can achieve a level of objectivity and reliability in performance evaluation that surpasses traditional human-based methods. However, despite these promising results, our study also uncovers a notable limitation of generative AI: the manifestation of human-like evaluative biases (Nisbett & Wilson, 1977). Specifically, the introduction of halo-effect information systematically influences GPT ratings, mirroring a common bias observed in human performance evaluations. This discovery not only underscores the nuanced complexities of integrating AI into management research but also prompts a critical examination of how such biases might be mitigated.

By highlighting the capabilities of LLMs to enhance the analytical tools available to organizational researchers, our study represents a key advancement in the integration of AI technologies in organizational research. This investigation not only contributes to the ongoing discussion on the integration of AI in organizational studies but also sets the stage for future research to build upon our findings, further refining and expanding the use of LLMs in theoretical and empirical inquiries. By leveraging LLMs, we illuminate a path forward for extracting nuanced concepts from text-based data, thereby refining and expanding the ways we measure constructs that are pivotal to management theory.

This approach not only enhances the precision of existing measures but also allows the identification and evaluation of previously unmeasurable constructs. For instance, consider the extraction of voice behaviors from meeting transcripts—a task that traditional methods might approach with significant difficulty due to its reliance on subjective interpretations or labor-intensive content analysis (Morrison, 2011). LLMs could dissect these transcripts to accurately identify instances of voice, offering a more precise and objective measure than ever before achievable. Through this inquiry, we pave the way for a novel wave of research that integrates cutting-edge computational techniques with the diverse organizational phenomena.

## The Literature Review and Hypothesis

**Theoretical Concept and Its Measurement in Organizational Research**

In organizational research, key concepts such as leadership, performance, and creativity form the bedrock for building and testing theories (Mathieu, 2016; Whetten, 1989). These concepts are often quantified through psychometric scales or proxy objective data to encapsulate the constructs central to our models and hypotheses (Edwards, 2001; Hinkin, 1995). However, traditional methods of measurement come with limitations, particularly the inability to capture spontaneous, real-time expressions of creativity and performance (Amabile & Pratt, 2016; Fisher & To, 2012). For instance, creativity, defined by the generation of ideas that are both novel and useful, frequently remains an untapped resource in research, with actual ideas seldom making it into empirical analysis (Berg, 2019; George, 2007; Zhou, Wang, Song, & Wu, 2017). Performance, representing the tangible contributions individuals make to their workplace, for knowledge workers, is predominantly manifested through knowledge-based outputs such as ideas, reports, memos, strategic plans, and marketing campaigns—yet, these rich data sources have historically been underutilized in management studies.

Recent shifts in research focus, for instance, within the domains of creativity and voice, have begun to address limitations in performance evaluation by concentrating on specific outputs rather than aggregated perceptions (Alvesson, 2001; George et al., 2014). This approach significantly broadens the theoretical landscape of performance assessment by leveraging language descriptions to encapsulate complex constructs. Human raters have traditionally served as the primary evaluators of such constructs (Viswesvaran & Ones, 2000). For instance, Berg's (2016) study, which involved using over 10,000 humans to assess idea novelty, exemplifies the extensive efforts to evaluate creative outputs objectively. Similarly, research on voice behaviors has highlighted the importance of specific voice content in organizational settings (Maynes & Podsakoff, 2014). These examples of creativity and voice are special forms of performance that demonstrate a key shortcoming in the literature: a vast array of potentially insightful data remains largely untapped and unevaluated (George et al., 2014; Tonidandel et al., 2018). By focusing on these specific outputs, researchers can gain a more precise understanding of performance, addressing the nuanced contributions of knowledge workers that traditional evaluations often miss (Spector & Pindek, 2016).

The advantage of focusing on language-based data lies in their ability to provide a more refined and precise measurement of constructs that are otherwise challenging to quantify (Short et al., 2018). Such an approach not only captures the content directly related to key concepts but also enables the exploration of constructs that are difficult to assess through conventional methods. Despite the availability of data recorded by digital tools and big data technologies, the capacity of humans to process and analyze this information remains constrained by time, resources, and cognitive limitations (George et al., 2014). Thus, this shift towards leveraging unobtrusive data sources aligns with the broader trends in management research, where there is an increasing emphasis on using digital footprints and naturally occurring data to gain insights into organizational phenomena (Hill, White, & Wallace, 2014; Knight, 2018). This approach can significantly expand the ways we measure constructs, offering a more comprehensive and accurate depiction of performance and behavior in organizational settings (Lazer et al., 2009).

**Review of Traditional NLP in Organizational Research**

The exploration of NLP and its applications within organizational research has introduced new methodologies for analyzing the vast quantities of textual data generated by and within organizations (Kobayashi, Mol, Berkers, Kismihók, & Den Hartog, 2018). Techniques such as Latent Dirichlet Allocation (LDA) for topic modeling, machine learning (ML) for text classification, and sentiment analysis have been widely adopted, each offering unique insights into organizational phenomena (Blei, Ng, & Jordan, 2003; Hannigan et al., 2019; Liu, 2020). For instance, sentiment analysis, often used to gauge the emotional valence of individuals, categorizes expressions as either positive or negative (Ashkanasy & Dorris, 2017). However, this simplistic classification fails to capture the nuanced spectrum of human emotions, a critical aspect in developing emotion-based theories within workplaces (Ashkanasy, Humphrey, & Huy, 2017). Traditional sentiment analysis is limited in its ability to measure the intensity and complexity of emotions. For example, feelings of anger and sadness can both be intense, yet they signify very different emotional states and require more precise measures to capture the richness of theoretically relevant constructs(Russell, 2003). Consequently, the inadequacy of traditional NLP techniques in capturing these nuances highlights the need for more advanced approaches that align with the detailed dimensions proposed in management theories.

Similarly, LDA and other topic modeling approaches have enabled researchers to distill thematic patterns from large text corpora, such as corporate documents or online forums (Corritore, Goldberg, & Srivastava, 2020). However, these methods primarily illuminate the presence of topics without providing deeper insights into the underlying reasons for their association or distinguishing the specific ways in which they differ (Grimmer & Stewart, 2013). This limitation constrains the depth of theoretical insights that can be extracted, hindering the development of theories that require a nuanced understanding of textual content (Schmiedel, Mu, & Brocke, 2018).

Moreover, while machine learning techniques have shown promise in extracting more complex and abstract theoretical constructs, such as personality traits, the necessity for extensive data labeling and validation processes has limited their practical application (Harrison et al., 2019; Kern, Rogge, & Howlett, 2019). The validity of these ML-based approaches often remains a subject of debate, presenting a barrier to their widespread adoption in organizational research (Tonidandel et al., 2018). Consequently, the potential of NLP in analyzing textual data in management studies faces significant limitations, highlighting a tension between the abundance of rich data and the current methodological toolkit's ability to fully harness it for theory development.

**LLMs in Organizational Research**

Unlike traditional AI systems that relied heavily on specific, pre-defined tasks and structured datasets, large language models (LLMs) such as GPT-4 represent a significant advancement. These generative AI models can engage in tasks without explicit prior training—a phenomenon known as zero-shot learning (Brown et al., 2020). For instance, when tasked with extracting creative performance from text data, these models can discern relevant information and provide ratings based on clearly defined criteria for creativity, mirroring human judgment (Bommasani et al., 2021).

In organizational research, human raters and experts have traditionally been integral to research processes (Duriau, Reger, & Pfarrer, 2007). They evaluate and code content to understand various constructs. However, LLMs show significant potential to supplement human raters due to their emerging human-like abilities. These abilities are evident in two main ways: First, LLMs can act without extensive pre-training, requiring only clear definitions and explanations, akin to the instructions given to human raters such as research assistants (Wei et al., 2022). This reliance on prompts ensures that LLMs can adapt to various tasks with minimal setup, enhancing their validity and flexibility in research applications. Second, LLMs introduce a level of randomness in their responses, similar to the variability seen in human judgments. This randomness can be adjusted through the "temperature" setting of the model (Holtzman, Buys, Du, Forbes, & Choi, 2020). A higher temperature results in more varied and creative responses, while a lower temperature yields more deterministic and focused outputs. This adjustability allows researchers to fine-tune the balance between consistency and variability, aligning the model's behavior with specific research needs.

The core human-like capabilities of LLMs make them promising tools in organizational research. For example, they can perform tasks such as evaluating the novelty and usefulness of ideas, which is the focus of our study. For example, a recent study exploring the use of LLMs as substitutes for human participants in marketing research found that LLM-generated outputs closely matched those from human surveys, with agreement rates exceeding 75% for both brand similarity measures and product attribute ratings (Li, Castelo, Katona, & Sarvary, 2024).

Therefore, we expect that an LLM can assess a task output like product naming for its novelty and usefulness by referencing defined criteria, providing insights directly relevant to theoretical constructs of creativity and effectiveness (Berg, 2019). While the potential of LLMs in these applications is promising, their viability as reliable tools for performance evaluation in organizational research requires thorough examination against well-established methods. Specifically, we need to compare LLM-based evaluations with those conducted by human evaluators, particularly multiple human raters who serve as proxies for ground truth (Amabile, 1982). This comprehensive examination will assess the extent to which LLM evaluations are consistent with human judgments, highlighting their strengths and identifying any weaknesses. In the following section, we develop specific hypotheses to systematically evaluate the abilities of LLMs in this domain.

**LLMs and Performance Evaluation Accuracy**

The potential capabilities of LLMs like GPT-4 in generating human-like ratings stem from their extensive training on vast corpuses comprising trillions of words (Brown et al., 2020). This extensive training endows these models with a knowledge base that far exceeds that of any single expert or group of human raters, equipping them with the ability to grasp a broad spectrum of complex concepts with a level of nuance and depth that traditional methods may not match (Bommasani et al., 2021). Empirical evidence from various tests, such as writing assessments and idea generation challenges, suggests that LLMs can effectively understand and apply evaluation criteria similar to human evaluators (Girotra, Meincke, Terwiesch, & Ulrich, 2023; Wei et al., 2022).

When provided with explicit standards of evaluation, LLMs can accurately comprehend both the criteria and the task outputs (Ouyang et al., 2022). This allows them to conduct evaluations that reflect a deep understanding of what constitutes good performance versus poor performance. Because LLMs are pre-trained on human language data, they are able to interpret definitions and standards effectively, leading to evaluations that should be comparable to those provided by human raters (Raffel et al., 2020; Yin et al., 2022). Therefore, we propose:

*Hypothesis 1 (H1): Evaluations generated by LLMs will be significantly related to evaluations conducted by human raters, demonstrating comparable understanding and application of evaluation criteria.*

Beyond their capability to match human evaluators in understanding criteria, LLMs also offer strengths that may lead to more reliable and consistent assessments. Unlike human raters, LLMs are not subject to fatigue, mood variations, or other idiosyncratic biases that can affect judgment(Bernardin & Buckley, 1981; Heilman, 2012; Kahneman, Sibony, & Cass.R.Sunstein, 2022) . LLMs can assess each output solely against the defined standards without the influence of previous judgments or personal biases (Brown et al., 2020; Yin et al., 2022). Research in performance rating has highlighted that a significant portion of variance in human ratings is due to individual idiosyncratic differences, with low to moderate correlation observed between raters (Hoffman, Lance, Bynum, & Gentry, 2010; O'Neill, Mclarnon, & Carswell, 2015).

For example, in a meta-analysis, Conway and Huffcutt (1997) reported moderate correlations between different rater sources, with supervisor-peer correlations being the highest ($\rho = .34$), followed by supervisor-self ($\rho = .22$) and peer-self ($\rho = .19$) correlations.

The consistency of LLMs means that, when constructs are well-defined, LLMs can provide more reliable assessments. They are less subject to individual errors and inconsistencies compared to human raters (Zhao et al., 2023). As a result, LLMs are likely to offer more consistent and reliable evaluations, capturing the intended constructs with higher fidelity (Khashabi, Kordi, & Hajishirzi, 2022; Wei et al., 2022). Although it might seem obvious that LLMs would be consistent with the same prompt, it's important to note that each response generated by an LLM is independent and can vary due to the inherent randomness in the model's output. This randomness can be controlled to some extent by adjusting the "temperature" setting, which influences the variability of the responses (Holtzman et al., 2020). Therefore, even with the same prompt, LLMs can produce different outputs, making their consistent performance noteworthy. Thus, we propose:

*Hypothesis 2 (H2): Evaluations generated by LLMs will demonstrate a higher degree of consistency and reliability in rating performance outcomes compared to evaluations conducted by individual human raters, showing less susceptibility to idiosyncratic biases.*

Moreover, the inherent randomness programmed into LLMs' algorithms introduces a level of variability in ratings, which could paradoxically enhance their utility. This built-in variance allows LLMs to simulate a range of perspectives, akin to multiple human raters independently evaluating the same output (Binz & Schulz, 2023; Brown et al., 2020; Dillion, Tandon, Gu, & Gray, 2023). Individual human raters often exhibit variance in their assessments (DeNisi & Murphy, 2017; Landy & Farr, 1980); however, their independent evaluations of the same object tend to offset random and idiosyncratic variances, converging towards a more accurate and true valuation of the concept at hand, serving as a proxy for ground truth (Hong & Page, 2004; Larrick & Soll, 2006; Scullen et al., 2000). Specifically, the randomness embedded within LLMs allows for diverse evaluations on the same text output, effectively mimicking the collective judgment of multiple human raters (Dillion et al., 2023). Consequently, when multiple GPT-generated ratings of the same output are aggregated, they tend to converge towards a more accurate reflection of the evaluated concept's true value (Larrick & Soll, 2006). This aggregation of multiple GPT ratings for the same performance reduces the impact of each GPT's biases and measurement errors, aligning more closely with the consensus ratings of multiple human experts.

*Hypothesis 3 (H3): Aggregated LLM evaluations will more closely align with the consensus ratings of multiple human experts, serving as a more accurate proxy for the ground truths of the evaluated concepts.*

**LLMs and Performance Evaluation Biases**

Our results suggest that GPT ratings are comparable to human ratings but exhibit higher consistency and reliability. Despite these advantages, it is crucial to examine whether LLMs are susceptible to the same cognitive biases that affect human raters. Human raters are inherently susceptible to a variety of cognitive biases in performance evaluations, including but not limited to the halo effect, leniency or severity bias, and confirmation bias (Landy & Farr, 1980). Among these, the halo effect stands out as a fundamental and pervasive bias in performance appraisals(Fisicaro & Lance, 1990; Nisbett & Wilson, 1977). For this study, we specifically focus on the halo effect due to its significant impact on performance evaluations.

The halo effect occurs when an observer's overall impression of a person, object, or entity influences their evaluation of specific attributes, leading to unduly favorable or unfavorable assessments (Nisbett & Wilson, 1977). This bias is prevalent in performance evaluations, where evaluators' general impressions of an employee can overshadow their objective assessment of specific performance dimensions (Balzer & Sulsky, 1992).

The halo effect is particularly significant due to its pervasive and unconscious nature (Nisbett & Wilson, 1977). It can cause raters to overlook specific weaknesses or strengths, resulting in evaluations that reflect general impressions rather than detailed, attribute-specific assessments (Murphy, Jako, & Anhalt, 1993; Thorndike, 1920). This bias systematically distorts performance appraisals across various contexts, from employee evaluations to consumer choices (Asch, 1946; Landy & Farr, 1980). Its impact on decision-making can lead to unfair or inaccurate assessments (Fisicaro & Lance, 1990), and it often resists correction even when raters are aware of its influence (Wetzel, Wilson, & Kort, 1981). These characteristics make the halo effect a crucial aspect to investigate in the context of LLM-generated evaluations, as understanding its potential presence in AI systems could provide valuable insights into the broader implications of cognitive biases in automated decision-making processes(Shrestha et al., 2019).

LLMs, by contrast, evaluate based on predefined criteria and are designed to assess text-based outputs independently of prior knowledge or impressions of the individual being evaluated (Bommasani et al., 2021; Vaswani et al., 2017). This objective, criteria-based approach should, in theory, make LLMs less prone to the halo effect. Even when presented with background information intended to introduce halo biases, LLMs process this information differently from humans (Wei et al., 2022). They do not form holistic impressions of individuals; instead, they analyze and rate each attribute based on the content and context of the provided text, thereby minimizing the risk of one attribute disproportionately influencing the overall evaluation(Brown et al., 2020; Gilardi, Alizadeh, & Kubli, 2023).

From a theoretical perspective, the absence of cognitive shortcuts in LLMs—shortcuts that humans often rely on—should reduce the susceptibility to the halo effect (Tversky &

Kahneman, 1974). Cognitive psychology posits that humans use heuristics to simplify decision-making processes, which can lead to systematic biases like the halo effect (Luan, Reb, & Gigerenzer, 2019). LLMs, however, process information systematically and adhere to the criteria set forth for evaluation, without the influence of extraneous factors such as an evaluator's mood, previous interactions with the subject, or irrelevant contextual information(Brown et al., 2020).

Additionally, LLMs' processing algorithms are designed to focus on specific prompts and tasks. When provided with explicit evaluation criteria, LLMs consistently apply these criteria to the text they analyze, ensuring that their assessments are based on the predefined standards rather than the overall impression introduced by background information(Bommasani et al., 2021). This systematic approach helps mitigate the halo effect because the model does not weigh background information as heavily as humans might. Instead, it evaluates each piece of information within the context of its relevance to the specific criteria being assessed.

*Hypothesis 4 (H4): LLM-generated evaluations will exhibit significantly lower susceptibility to the halo effect in performance assessments compared to evaluations conducted by human raters.*

In addition to the baseline hypotheses, our study recognizes the potential for LLMs to offer more nuanced insights into performance evaluations, warranting further exploration through specific research questions. Without developing a priori assumptions, we seek to explore several critical aspects of LLM evaluations. First, we investigate how the number of LLM-generated ratings impacts the accuracy of performance evaluations and how this relationship is moderated by the temperature setting used during evaluation. This focus aims to understand how varying the number of evaluations and adjusting the model's randomness can mimic the diverse perspectives and collective wisdom of human experts (Hong & Page, 2004). Additionally, we examine the extent to which LLM evaluations reflect contextual biases present in the text, such as framing effects introduced by additional context about the evaluated individual. This exploration will help us refine LLM applications and mitigate unintended biases. Furthermore, we assess how well LLMs distinguish between different evaluative dimensions, such as overall performance, novelty, and usefulness dimensions, compared to human raters, providing insights into their nuanced understanding and evaluation capabilities.

## Methods

### Overview Design

Our research uses two distinct datasets to examine and compare the evaluative precision of LLMs against human judgment across varied environments. The first study uses performance outputs from a controlled laboratory setting, where participants, including professionals and students, completed professional tasks. These performance outputs were subsequently evaluated by both human raters and LLMs. This setup allows us to compare LLM evaluations to human evaluations against various benchmarks, providing a controlled environment to assess the accuracy and reliability of LLM-generated ratings.

To achieve this comparison, we employed both descriptive and meta-analytical approaches. The descriptive analysis examined the correlation between GPT ratings and human ratings to determine their comparability. Additionally, we conducted a meta-analysis using Hunter and Schmidt (2004) methods to further validate the comparability of GPT evaluations with human raters. This approach allowed us to estimate the true correlation between raters and its variability, providing a robust statistical foundation for our comparisons.

The second study transitions to a field setting at a prominent taxi company in China, providing an ecologically valid examination of real-world performance evaluations within the context of a promotional examination. In this study, we replicate our findings using text outputs from a real organizational setting. We also examine potential biases by manipulating the background information of the employees being evaluated to assess the extent to which LLMs exhibit the halo effect. Both studies aim to provide a comprehensive analysis of how LLMs perform relative to human evaluators across different performance tasks, elucidating the strengths and potential limitations of LLMs in performance evaluation.

**Study 1**

**Sample and Procedurals**

The first study used performance outputs generated by 130 participants who took part in a laboratory setting to complete professional tasks. The sample included working professionals as well as college students. Each participant completed four tasks, resulting in a total of 520 textual performance outputs. The tasks were designed to assess different professional skills: writing a job search cover letter to evaluate persuasive writing capabilities, creating and justifying new product names to gauge creative thinking, developing solutions for team conflict scenarios to assess interpersonal and conflict resolution skills, and designing an AI-integrated college course curriculum to test innovative thinking in educational contexts. Detailed task descriptions can be found in the online appendix[1].

The text-based performance outcomes of these tasks were evaluated on three dimensions: overall quality, novelty, and usefulness (Montag, Maertz Jr., & Baer, 2012), each rated on a scale from one to ten. In later descriptions, we refer to these dimensions as *Overall*, *Novelty*, and *Usefulness*. An online panel of independent judges, each blind to the study's hypotheses and the conditions under which each task output was created, assessed each performance output. Prior to evaluation, the raters were provided brief instructions to clarify the evaluation criteria. Each work was rated by six raters, with a total of 486 human raters initially participating in the evaluation process. Raters who spent an unreasonably short amount of time on the tasks were removed, resulting in a final pool of 382 valid raters. On average, each valid rater provided ratings for approximately 8 performance outputs. The tasks being rated were presented in a random sequence to each rater, which helped eliminate figure, sequence, and other potential confounding factors in the evaluations.

---

[1] All data, code, supplementary materials, and appendices related to this research are accessible via the Open Science Framework (OSF). You can find these resources at the following link https://osf.io/png9e/?view_only=21043a2f38034feb87e9c70f7265daad

To evaluate the internal consistency of the ratings, we calculated Cronbach's alpha for each dimension, which resulted in values of .72 for overall quality, .71 for novelty, and .69 for usefulness. Cronbach's alpha was chosen because it is a widely used measure of internal consistency, indicating how closely related a set of items are as a group.

For evaluating performance using LLMs, we employed the OpenAI API (GPT-4 Model, OpenAI, 2023), programmed in Python. Each response was independently rated six times by GPT-4. To ensure a fair comparison, GPT-4 was provided with the same clear evaluation criteria in their prompts as given to the human judges. This method involves no prior training on specific tasks, ensuring that each evaluation adheres to predefined standards similar to those applied in human assessments. Unlike traditional NLP methods that generate deterministic ratings, GPT ratings introduce an element of variability akin to human judgment, where identical inputs might produce varied ratings. This variability is regulated by the "temperature" parameter in the model, which can be adjusted between 0 and 2; a higher temperature increases randomness. For our comparative analysis between GPT and human ratings, we maintained the temperature setting at the default value of 1. Further investigations were conducted to assess the effects of this inherent randomness on the consistency and accuracy of GPT ratings. Our analysis revealed high reliability of GPT ratings across the three dimensions, with Cronbach's alpha values of .93 for overall quality, .93 for novelty, and .90 for usefulness.

**Results**

   ***Comparable GPT Ratings to Human Ratings.*** Our analysis begins by examining the correlation between single GPT ratings and single human ratings to determine their comparability. The descriptive analysis in Figure 1 shows that the correlation between a single GPT rating and a single human rating (noted as GPT[1]-Human[1]) is .41 for Overall in Task 1. This correlation is comparable to the correlation between two human ratings (noted as Human[1]-Human[1]), which is .35 for Overall in Task 1. The results suggest that individual GPT ratings are as comparable to individual human ratings. In addition, the descriptive analysis in Figure 1 reveals that the correlations between two GPT ratings (noted as GPT[1]-GPT[1]) are higher than the correlations between two human ratings (Human[1]-Human[1]), implying that GPT ratings may have an advantage over human ratings in terms of evaluation consistency. For instance, in Task 1 for Overall, GPT(1)-GPT(1) shows a correlation of .72, which is higher than Human(1)-Human(1)'s .35. Moreover, Figure 1 also illustrates that the correlations between six GPT ratings and six human ratings (noted as GPT[6]-Human[6]) are substantially higher than the correlations for GPT(1)-Human(1), suggesting the potential for improving rating performance through aggregating multiple raters. For example, GPT(6)-Human(6) has a correlation of .70 compared to GPT(1)-Human(1)'s .41 in Task 1 for Overall.

   To further validate the comparability of GPT evaluations with human raters, we conducted a meta-analysis using Hunter and Schmidt (2004) methods to estimate the true correlation between raters and its variability. Each task output was evaluated by six human raters, allowing us to estimate the correlation between one human rater and another human rater (Human[1]-Human[1]). Similarly, each task was rated independently by GPT-4 six times, enabling us to estimate the correlation between one GPT rating and one human rating (GPT[1]-Human[1]). This approach allowed us to directly compare the comparability of GPT ratings with human ratings.

Each performance outcome was rated by six human individuals and by GPT-4 six times, generating multiple pairwise correlations. For each outcome, the ratings by human raters resulted in 15 unique correlations (Human[1]-Human[1]). Similarly, the ratings by GPT-4 produced 36 unique correlations with human raters (GPT[1]-Human[1]). These correlations were treated as individual data points in our meta-analysis. By aggregating these correlations across many performance outcomes, we conducted meta-analyses to compute an overall correlation coefficient, providing a more accurate and reliable estimate of the true score correlations. This meta-analytic approach synthesizes multiple correlation estimates, offering a generalized understanding of the comparability between GPT and human ratings.

The meta-analysis uses 95% confidence intervals (CIs) to evaluate the correlations. For instance, as Table 1 shows, in Task 1 for Overall, the 95% CI for the correlation coefficient between a single GPT rating and a single human rating (GPT[1]-Human[1]) is [.39, .44], which slightly overlaps with the 95% CI for the correlation between two human ratings (Human[1]-Human[1]), which is [.29, .40]. And in Task 4 for Novelty, GPT(1)-Human(1)'s 95% CI [.38, .43] overlaps with Human(1)-Human(1)'s 95% CI [.30, .42]. The overlap indicates that the comparability of GPT ratings is similar to that of human ratings.

Additionally, we compared the correlation of a single GPT rating with the average of five human ratings (noted as GPT[1]-Human[5]) to the correlation of a single human rating with the average of five human ratings (Human[1]-Human[5]). The average scores from five human judges (Human[5]) serve as a proxy for the ground truth, combining multiple independent evaluations to reduce individual biases and provide a more accurate benchmark.

In Figure 2, we can observe that the correlations for GPT(1)-Human(5) and Human(1)-Human(5) are comparable in most cases. For instance, in Task 1 for Overall, GPT(1)-Human(5) shows a correlation of .60, whereas Human(1)-Human(5) shows a correlation of .50. Similar trends are observed across other tasks, where GPT(1)-Human(5) often shows slightly higher or comparable correlation coefficients to Human(1)-Human(5). For example, in Task 2 for Usefulness, GPT(1)-Human(5) has a correlation of .42 compared to Human(1)-Human(5)'s .44, and in Task 3 for Overall, GPT(1)-Human(5) shows a correlation of .64 compared to Human(1)-Human(5)'s .55.

The meta-analysis confirms this observation, Table 2 indicates that the correlation between GPT(1) and Human(5) is comparable to, and sometimes higher than, the correlation between Human(1) and Human(5). For example, in Task 1 for Overall, the 95% CI for GPT(1)-Human(5) is [.58, .62], which slightly exceeds the 95% CI for Human(1)-Human(5), which is [.44, .57]. And in Task 2 for Usefulness, the 95% CI for GPT(1)-Human(5) is [.42, .47], which is notably higher than Human(1)-Human(5)'s [.19, .32].

These findings from both the descriptive analysis and the meta-analysis support Hypothesis 1, affirming that LLM-generated evaluations possess a degree of comparability in rating performance outcomes similar to human ratings. By matching the evaluative precision traditionally expected from human raters, GPT ratings demonstrate their potential as reliable substitutes for human evaluations in performance assessment contexts.

***Higher Consistency and Reliability of GPT Ratings***. Our analysis reveals a clear pattern of higher consistency and reliability in single GPT ratings compared to single human ratings across various tasks and scoring dimensions. Figure 1 illustrates these findings, showing that in Task 1, the correlation coefficients for single GPT ratings (noted as GPT[1]-GPT[1]) stand at .72 for Overall, .56 for Novelty, and .69 for Usefulness. These values significantly outperform the correlations for single human ratings, which are .35 for Overall, .32 for Novelty, and .33 for Usefulness. Similar patterns of GPT's superior performance are observed in Tasks 2, 3, and 4, demonstrating a consistent trend of higher reliability in GPT evaluations.

Building on these descriptive findings, the meta-analysis results presented in Table 1 further validate the superior consistency of GPT ratings. The meta-analysis employs 95% confidence intervals (CIs) to assess the statistical significance and reliability of the correlations. For instance, in Task 1 for Overall, the 95% CI for the correlation coefficient for GPT(1)-GPT(1) is [.70, .74], which notably surpasses the [.29, .40] observed for Human(1)-Human(1). This trend of enhanced consistency is evident across Novelty and Usefulness dimensions, where GPT ratings achieve correlations with 95% CIs of [.53, .59] and [.67, .71], respectively, indicating significant improvements over the lower correlations noted in human ratings, which are [.26, .39] for Novelty and [.29, .37] for Usefulness.

These findings suggest that GPT ratings exhibit lower variability and thus provide more consistent evaluations compared to individual human raters. This enhanced consistency and reliability support Hypothesis 2, affirming that GPT-generated evaluations demonstrate a higher degree of consistency and reliability in rating performance outcomes compared to human ratings.

***Aggregating of multiple GPTs***. Both Figure 3 and Table 3 collectively demonstrate that while increasing the number of evaluators generally enhances correlation scores for both human and GPT ratings, GPT consistently achieves higher correlations at each level. This trend indicates that for both GPT and human, idiosyncratic variances can be offset by increasing the number of raters. These findings underscore GPT's capability to minimize individual idiosyncratic errors and enhance reliability, making it particularly suited for scenarios demanding high precision and consistent evaluations.

Figure 4 illustrates that GPT(6) achieves a reliability measure that adheres to established standards, where correlations greater than .5 are considered strong and those above .7 as very strong (Cohen, 1988; Hemphill, 2003). For example, in Task 1, GPT(6) shows high correlations with the ground truths for Overall, Novelty, and Usefulness at .70, .60, and .69, respectively. These strong correlations indicate that aggregated GPT ratings align closely with the proxy ground truths provided by multiple human raters(Hunter & Schmidt, 2004; Nunnally, 1978).

Furthermore, GPT ratings exhibit lower idiosyncratic variance, leading to diminishing returns when increasing the number of GPT raters compared to human raters. The meta-analysis in Table A1 of the Appendix demonstrates that fewer GPT raters are required to achieve an acceptable level of agreement compared to human raters. For instance, while the 95% CI for Human(1)-GPT(6) starts at [.41, .53], increasing human raters to five enhances this range to [.65, .72]. Conversely, starting from a higher baseline, a single GPT rater's 95% CI of [.57, .66] quickly extends to [.64, .69] with just one additional GPT rater, indicating a more rapid convergence towards higher reliability.

Overall, these results provide support for Hypothesis 3. The evidence demonstrates that aggregated GPT evaluations more closely align with the consensus ratings of multiple human experts, serving as a more accurate proxy for the ground truths of the evaluated concepts. This underscores the effectiveness of GPT ratings in producing reliable and precise evaluations, confirming their potential as robust alternatives to traditional human assessments.

**Additional Analyses**

***The discriminative validity of GPT ratings compared to human ratings***. To further understand the performance of GPT ratings, we conducted an analysis to evaluate their discriminative validity compared to human ratings. This analysis aimed to determine whether GPT raters could differentiate between score dimensions more effectively than human raters, which is crucial for ensuring nuanced and accurate evaluations in performance assessment.

We found that GPT demonstrates a distinct capability to differentiate between score dimensions more effectively than human raters. For each task, we calculated the correlations within each rater (single GPT or single human) across different score dimensions, namely O-N (Overall to Novelty), O-U (Overall to Usefulness), and N-U (Novelty to Usefulness). Meta-analysis was employed to synthesize these results.

As detailed in Figure 5 and supported by the data in Table A2 of the Appendix, GPT ratings consistently exhibit lower correlations between dimensions than human ratings across all tasks, indicating a higher discriminative capacity. For example, in Task 1, while human raters show a high correlation range between Overall and Usefulness (O-U) ([.89, .91]), GPT raters exhibit a lower correlation range ([.84, .88]). This pattern of lower correlations in GPT ratings is also evident in Task 2, where the correlation between Novelty and Usefulness (N-U) is notably lower for GPT ([.21, .35]) compared to humans ([.52, .65]). Similarly, in Task 3 and Task 4, GPT continues to demonstrate more variability and lower correlations between different dimensions (e.g., O-U in Task 4 for GPT [.68, .77] vs. Human [.83, .89]). These findings suggest that GPT ratings can better distinguish between the meanings of different score dimensions.

***Trade-off Between Individual Variance and Group Accuracy in GPT Ratings***. To explore the implications of using multiple GPT raters, we investigated the relationship between individual variance and the accuracy of GPT ratings. This analysis aimed to understand how the variability introduced by different temperature settings and the number of raters affects the overall accuracy of evaluations, which is essential for leveraging the "wisdom of the crowd" principle in performance assessments (Surowiecki, 2004).

Prior research suggests that a diverse group of independent judges typically yields more accurate ratings (Hong & Page, 2004). Diversity implies greater variance at the individual level. To investigate the relationship between individual variance and GPT rating accuracy, we manipulated the temperature parameter of GPT to introduce varying levels of randomness.

Figure 6 illustrates the accuracy of GPT ratings across different tasks, scoring dimensions, rater sizes, and temperature settings, using Human(6) as the proxy for ground truth. The results reveal that GPT rating accuracy increases with the number of raters, although each additional rater contributes progressively smaller gains. Temperature also significantly impacts accuracy; notably, lower temperatures (e.g., .05, .25) enhance accuracy in smaller rater groups (e.g., 1, 2), while higher temperatures (e.g., .50, .75, 1.00) prove more effective in larger rater groups (e.g., 5, 6). Moreover, the optimal temperature varies, indicating a nuanced trade-off between randomness and accuracy across different tasks and score dimensions.

Regression models further confirm this dynamic. Table 4 shows that Rater Size positively influences GPT rating accuracy ($b = .035$, $p < .001$ for Overall; $b = .039$, $p < .001$ for Novelty; $b = .050$, $p < .001$ for Usefulness), suggesting that increasing the number of raters improves accuracy. However, the diminishing returns from additional raters are evident, as indicated by the negative coefficients of the squared term Rater Size*Rater Size ($b = -.004$, $p < .001$ for Overall;

b = -.005, p < .01 for Novelty; b = -.006, p < .001 for Usefulness). The relationship between temperature and accuracy is significantly negative (b = -.096, p < .001 for Overall; b = -.061, p < .01 for Novelty; b = -.039, p < .05 for Usefulness), yet positively moderated by Rater Size (b = .021, p < .001 for Overall; b = .018, p < .01 for Novelty; b = .018, p < .001 for Usefulness). This suggests that increasing the number of raters can not only mitigate but actually reverse the adverse effects of higher temperature settings, leading to higher accuracy. High temperature settings require a larger number of raters to achieve more accurate evaluations, effectively balancing individual variance and group accuracy.

## Study 2

### Sample and Procedurals

The performance outputs in Study 2 were from an internal promotion selection test conducted by a prominent taxi company in China. As part of the selection process for middle management positions, 112 employees were tasked with responding to two out of fourteen promotion tasks. These tasks, designed by the company's management team, aimed to address prevalent business management challenges, such as identifying reform opportunities for significant issues within the taxi industry, enhancing corporate branding and image, and advancing the digital transformation of the enterprise.

This dataset comprised 224 textual responses and served as the basis for comparing performance evaluations within a practical setting. Similar to Study 1, each response was rated by six GPT and six human raters across three score dimensions: Overall, Novelty, and Usefulness (for detailed task descriptions, human ratings, and GPT ratings, refer to Supplementary Materials). We recruited a total of 420 raters participating in the evaluation.

Study 2 aimed to replicate the findings of Study 1 in a field setting while also examining prevalent biases in performance evaluations, particularly focusing on halo effects. The real organizational context provided a rich environment for manipulating the employees' prior backgrounds, such as education and prior performance, to create different halo conditions.

Specifically, we provided raters with background information about the target employees alongside their performance outputs. This background information was carefully crafted to be theoretically independent of the actual performance outcomes, allowing us to isolate its influence on ratings. We created three variations of this background information: a positive version that presented strongly favorable details about the employee's education, work experience, and past performance; a neutral version that offered a balanced view of the employee's background; and a negative version that included critical information about the employee's history. While maintaining the same underlying content across all versions, we subtly altered the phrasing and specific details, such as the prestige of the employee's alma mater, to create distinct impressions. Our goal was to assess whether the presence of this extraneous background information would sway raters' evaluations of the performance outputs, which ideally should be judged solely on their merits against the established criteria.

We created three conditions for both human and GPT ratings. Figure 7 illustrates our research design of Study 2. The halo condition consisted of performance outputs randomly assigned one of the halo conditions, with roughly equal numbers receiving positive, neutral, and negative halos (37/38/37). Raters (both human and GPT) were provided both the halo text and the performance output text. The halo mitigation group maintained the same halo distribution but included instructions to the raters to not consider the background information when they rate performance. In the control group, raters were provided only the performance output text for evaluation, without any halo background information[2].

**Results**

***Replication of Study 1 Findings.*** Utilizing the dataset from Study 2, we validated the primary findings of Study 1 (for detailed results, please see Appendix[3]). First, GPT ratings were comparable to human ratings. Comparing single GPT ratings with single human ratings (GPT[1]-Human[1]) and the correlations between two human ratings (Human[1]-Human[1]), the results showed that GPT evaluations matched the evaluative precision of human raters. For instance, Figure A1 of the Appendix showed that, for Overall, GPT(1)-Human(1) had a correlation of .23, and Human(1)-Human(1) had a correlation of .22. The 95% confidence intervals (CIs) for these correlations confirmed the alignment of GPT ratings with human ratings, demonstrating their comparability. For example, for Overall, the 95% CI for GPT(1)-Human(1) was [.21, .25], overlapping with Human(1)-Human(1)'s [.19, .25].

Second, GPT ratings exhibited higher consistency and reliability than individual human ratings. The correlation between one GPT rating and another GPT rating (GPT[1]-GPT[1]) was higher than the correlation between one human rating and another human rating (Human[1]-Human[1]), indicating lower idiosyncratic variance in GPT ratings. For instance, for Overall, the correlation for GPT(1)-GPT(1) was .38, which was higher than Human(1)-Human(1)'s .22.

---

[2] For human ratings, the Cronbach's alpha values in the control group are 0.631, 0.540, and 0.644 for *Overall*, *Novelty*, and *Usefulness* respectively. In the halo group, these values increase to 0.714, 0.678, and 0.651. For the halo mitigation group, the values slightly decrease to 0.643, 0.604, and 0.579. Conversely, for GPT ratings, the control group values are notably higher at 0.780, 0.754, and 0.715 for *Overall*, *Novelty*, and *Usefulness* respectively. The halo group shows even greater internal consistency with values of 0.909, 0.891, and 0.896, while the halo mitigation group maintains high reliability with values of 0.855, 0.852, and 0.841. Overall, both human and GPT ratings exhibit relatively high reliability, with GPT ratings consistently surpassing those of human ratings. This finding aligns closely with the results from Study 1, further validating the robustness of GPT in providing consistent and reliable evaluations.

[3] To distinguish the four tasks of Study 1, we denote the task of Study 2 as Task 5 in the Appendix.

Additionally, comparing single GPT ratings to the average of five human ratings (GPT[1]-Human[5]) indicated considerable alignment, further validating GPT's consistency and accuracy (.37 for Overall, .39 for Novelty, .32 for Usefulness).

Third, aggregated GPT ratings closely correlated with multiple human ratings, reinforcing their reliability as an alternative to human raters. GPT ratings showed strong correlations to the proxy of ground truths (e.g., Human[6]). As detailed in Figure A1 of the Appendix, the correlation coefficients for GPT(6)-Human(6) are .57 for Overall, .60 for Novelty, and .52 for Usefulness.

Finally, we observed the interaction between temperature settings and rater size, revealing a trade-off between randomness and overall accuracy. As shown in Table A8 in the Appendix, regression models affirmed the robustness of our initial findings, demonstrating that increasing the number of raters improved accuracy ($b = .039$, $p < .001$ for Overall; $b = .045$, $p < .001$ for Novelty; $b = .052$, $p < .001$ for Usefulness), though gains diminished with each additional rater ($b = -.005$, $p < .001$ for Overall; $b = -.005$, $p < .001$ for Novelty; $b = -.006$, $p < .001$ for Usefulness). Lower temperature settings enhanced accuracy in smaller rater groups, while higher settings were more effective in larger groups ($b = .021$, $p < .001$ for Overall; $b = .020$, $p < .001$ for Novelty; $b = .018$, $p < .001$ for Usefulness). This trade-off indicates that while randomness can reduce individual rating validity, it enhances overall validity when multiple ratings are aggregated.

***Results for the halo effect.*** We begin by presenting the descriptive results. As stated previously, multiple human ratings, such as Human(6), serve as a proxy for the ground truth. Using this as a benchmark, we observed that the correlation between Human(6) and GPT(6) without halo information is strong across the dimensions of Overall, Novelty, and Usefulness (.57/.60/.52, respectively).

Figure 8 shows that when halo information is introduced, both the GPT halo group (.34/.36/.38) and the GPT halo mitigation group (.39/.43/.41) exhibit reduced correlations with Human(6) across the dimensions of Overall, Novelty, and Usefulness, respectively. This indicates that the presence of a halo significantly impacts the accuracy of GPT ratings. The halo mitigation group, which included instructions to disregard the background information and rate the responses objectively, showed some reduction in the halo effect but did not eliminate it entirely.

Similarly, when human raters were presented with halo information, both the human halo group (.43/.49/.43) and the human halo mitigation group (.42/.45/.44) also demonstrated diminished correlations with GPT(6). This suggests that human ratings are affected by the halo effect, which explicit instructions alone cannot fully mitigate.

Interestingly, the correlation between Human(6) with halo and GPT(6) with halo (.67/.64/.65) is higher than the benchmark correlation of Human(6)-GPT(6). This may be because the halo induces a convergence in GPT and human ratings, leading to higher ratings for positive halo and lower ratings for negative halo, thereby making their ratings more similar.

To assess the influence of different halo types (positive, neutral, negative) on ratings, we conducted paired t-tests across various groups, detailed in Table 5. The analysis confirms the significant presence of the halo effect in both GPT and human ratings. A positive halo significantly increases scores in both GPT ratings (Overall: Mean = .32, $p < .001$; Novelty: Mean = .46, $p < .001$; Usefulness: Mean = .43, $p < .001$) and human ratings (Overall: Mean = .52, $p < .001$; Novelty: Mean = .51, $p < .001$; Usefulness: Mean = .38, $p < .001$). Conversely, a negative halo substantially lowers scores for GPT ratings (Overall: Mean = -1.90, $p < .001$; Novelty: Mean = -1.76, $p < .001$; Usefulness: Mean = -2.01, $p < .001$) and human ratings (Overall: Mean = -.86, $p < .001$; Novelty: Mean = -.89, $p < .001$; Usefulness: Mean = -.85, $p < .001$). Neutral halos did not significantly affect human ratings but led to reduced scores in GPT ratings (Overall: Mean = -.77, $p < .001$; Novelty: Mean = -.91, $p < .001$; Usefulness: Mean = -.75, $p < .001$).

In groups with halo mitigation instructions, the halo effect persisted for both GPT and human raters, albeit with diminished impacts. Positive halos still led to higher scores for both GPT (Overall: Mean = .18, $p < .01$; Novelty: Mean = .21, $p < .05$; Usefulness: Mean = .32, $p < .001$) and human ratings (Overall: Mean = .57, $p < .001$; Novelty: Mean = .36, $p < .01$; Usefulness: Mean = .41, $p < .001$). Negative halos continued to lower scores in both GPT (Overall: Mean = -1.26, $p < .001$; Novelty: Mean = -1.29, $p < .001$; Usefulness: Mean = -1.47, $p < .001$) and human ratings (Overall: Mean = -.46, $p < .001$; Novelty: Mean = -.32, $p < .05$; Usefulness: Mean = -.46, $p < .001$). Neutral halos did not significantly impact human ratings but still resulted in lower GPT scores (Overall: Mean = -.59, $p < .001$; Novelty: Mean = -.80, $p < .001$; Usefulness: Mean = -.65, $p < .001$).

This analysis demonstrates that GPT is surprisingly more susceptible to halo biases than humans. In particular, GPT rating is particularly susceptible to negative halos, significantly reducing scores under halo-induced conditions, with effects much stronger than those observed in human ratings. This heightened sensitivity to halo biases suggests that extraneous background information more strongly influences GPT's evaluations. This does not support Hypothesis 4, which predicted that LLM-generated evaluations would show significantly lower susceptibility to the halo effect in performance assessments compared to those conducted by human raters.

This finding highlights a potential vulnerability in LLM evaluations, especially in situations where contextual biases are common.

## Discussion

Across two studies, we demonstrated that large language models (LLMs) such as GPT-4 can provide performance evaluations that are highly comparable to those of human raters, often surpassing them in consistency and reliability. Specifically, aggregated GPT ratings showed strong correlations with multiple human ratings across various dimensions, including overall quality, novelty, and usefulness. The results indicated that LLMs offered an alternative method for achieving reliable and consistent performance evaluations. Moreover, while LLMs exhibited susceptibility to biases such as the halo effect, the introduction of mitigation instructions could reduce, though not eliminate, these biases. These findings highlight the potential of LLMs to enhance traditional human evaluations in organizational research, offering a powerful tool for analyzing complex textual data with greater accuracy and efficiency.

**Implications for Theory and Method**

Our study presents several significant implications for theory and method, highlighting the significant potential of LLMs in organizational research. First, our findings demonstrate that LLMs can significantly improve the accuracy, consistency, and reliability of performance evaluations. Traditionally, measuring constructs like creativity, performance, and usefulness has relied heavily on human judgment, which is often subject to biases and variability (Viswesvaran, Ones, & Schmidt, 1996). By integrating LLMs, researchers can refine and operationalize constructs with greater precision. The ability of LLMs to handle large volumes of text data and extract meaningful patterns allows for more nuanced and comprehensive empirical validation of theoretical models. This capability not only strengthens the validity of existing constructs but also opens up new avenues for exploring theoretical relationships in organizational behavior.

Second, our study demonstrates that LLMs offer a novel approach to measuring constructs that were previously difficult to capture (George et al., 2014). By showing that LLMs can reliably provide ratings for text-based outputs, we open the door for analyzing context-specific performance observations rather than general perceptions. This capability allows for the measurement of new constructs that reflect the evolving nature of organizational tasks and behaviors. For instance, constructs related to instant feedback and creative ideas can now be accurately measured and analyzed. The digital traces left by employees, such as emails, reports, and meeting notes, serve as rich data sources for LLMs to analyze, providing insights into performance that go beyond aggregated perceptions (Knight, 2018). This context specific evaluation ability facilitates the development of new theoretical models that incorporate dynamic data, offering a more immediate and nuanced understanding of organizational phenomena.

Third, our findings raise important questions about who holds the ground truth in performance evaluations. Traditionally, consensus human raters have been used as proxies for the ground truth due to their ability to understand and interpret complex constructs(Amabile, 1982). However, why should LLMs not also be considered as valid proxies? Human raters have long been the primary choice for a wide range of research processes, from scale development to evaluating outputs, due to their perceived ability to understand and interpret complex constructs. Traditional NLP methods have often been limited in their ability to assist in these tasks. Our study empirically compares human raters with LLMs, demonstrating that AI can perform evaluative tasks with a level of precision that rivals human raters. This challenges existing theories that rely on human judgment as the gold standard for performance evaluation and opens the door to new theoretical frameworks that integrate both human and AI-driven evaluations. By acknowledging the strengths and limitations of both, researchers can enhance the accuracy and depth of organizational research, leveraging the complementary capabilities of humans and AI for more comprehensive evaluations.

Fourth, one of the most significant methodological implications is the scalability of LLMs. Unlike human raters, who are limited by time and cognitive resources, LLMs can process vast amounts of text data quickly and efficiently. This scalability allows for large-scale studies that were previously impractical due to the limitations of human evaluators. For example, traditional NLP methods have been used to process email data with limited options for extracting meaningful constructs (Kleinbaum, Stuart, & Tushman, 2013). However, with LLMs, researchers can now analyze massive datasets, such as entire email corpora, meeting transcripts, and extensive reports, to develop new, meaningful constructs. This ability to scale up evaluations without compromising accuracy or reliability is a significant advancement, enabling more ambitious and wide-reaching research projects in organizational studies.

Related, LLMs enable a level of granular analysis that is difficult to achieve with traditional methods. Our study shows that LLMs can uncover subtle patterns and insights within textual data that might be overlooked by human raters. This granular analysis allows for a deeper understanding of complex constructs and interactions within organizational settings. This capability is particularly valuable for developing and testing nuanced theoretical models that require detailed and specific data inputs(Srivastava, Goldberg, Manian, & Potts, 2018).

**Implications for Practice**

Our study also has several implications that can impact organizational practices, particularly in the areas of performance evaluation, talent management, and decision-making.

First, integrating LLMs into performance evaluation processes can enhance the objectivity and consistency of assessments. Organizations often rely on supervisors to evaluate employee performance, which can be subject to biases and variability. By incorporating LLMs, organizations can achieve more consistent and reliable evaluations, reducing the impact of individual biases and improving the fairness of performance reviews. This can lead to more equitable decisions regarding promotions, bonuses, and career development opportunities, ultimately fostering a more meritocratic organizational culture.

Second, the ability of LLMs to analyze large volumes of text data efficiently makes them valuable tools for ongoing performance management and feedback. Traditional performance evaluations are typically conducted at fixed intervals, such as annually or semi-annually (Buckingham & Goodall, 2015). However, LLMs can provide real-time, continuous feedback by analyzing ideas, reports, and other knowledge outputs as they are produced. This allows managers to identify performance issues or exemplary behavior promptly, providing timely feedback that can enhance employee development and productivity.

Third, the findings of our study highlight the importance of addressing biases in AI-driven evaluations. While LLMs offer many advantages, they are not immune to biases such as the halo effect. Organizations must implement strategies to mitigate these biases, such as providing clear guidelines to LLMs and incorporating oversight mechanisms. Additionally, organizations need to preprocess information presented to AI by removing performance-irrelevant data to avoid biases. This ensures that AI-driven evaluations remain fair and accurate, maintaining trust in the system among employees and stakeholders.

Finally, the use of LLMs in organizational decision-making can enhance the depth and breadth of insights available to leaders. By leveraging the analytical capabilities of LLMs, organizations can gain a more nuanced understanding of employee performance, customer feedback, and market trends (Li et al., 2024). This can inform strategic decisions, such as identifying areas for improvement, developing new products or services, and optimizing resource allocation. The ability to analyze complex textual data at scale allows organizations to make more informed, data-driven decisions, ultimately driving better business outcomes.

**Limitations and Future Research Directions**

We acknowledge several limitations of this research. First, our focus was on performance evaluations based on textual data, which does not cover other critical forms of performance output, such as oral presentations or collaborative tasks. Future research should integrate multimodal data to enhance the comprehensiveness of AI-driven evaluations.

Another limitation of Study 2 is the lower inter-rater reliability, with coefficients below the conventional threshold of 0.70 (Lance, Butts, & Michels, 2006). This suggests a difficulty for humans to reach agreement on complex and ambiguous performance outcomes. Despite this, GPT ratings demonstrated notably higher reliability and strong correlations with aggregated human ratings, suggesting GPT's viability even in ambiguous situations. Future research should explore the boundary conditions across various tasks and constructs to thoroughly examine the validity of using LLMs in evaluating human knowledge work.

Future research can pursue several promising directions. One area is the application of LLMs in evaluating team dynamics and collaboration. By analyzing communication patterns and collaborative outputs, researchers can assess team performance, identify areas for improvement, and provide feedback to enhance team effectiveness(Kozlowski & Chao, 2018; Waller & Kaplan, 2018). Another area is the analysis of employee voice and feedback, where LLMs can evaluate the content and tone of feedback, providing insights into employee sentiment and areas of concern (Maynes & Podsakoff, 2014). This can help develop strategies to improve employee engagement and satisfaction.

Ongoing research should focus on the ethical implications and fairness of AI-driven evaluations. It is crucial to minimize biases and ensure equitable treatment across diverse employee groups to maintain trust and integrity in AI applications within organizations (Kleinberg & Ludwig, 2019). AI has the potential to reduce adverse impacts and disparities from human biases in performance evaluations (Bobko & Roth, 2013). By rigorously testing and refining AI systems, organizations can work towards reducing evaluation gaps between demographic groups, promoting a more inclusive and fair workplace (Hekman, Aquino, Owens, Mitchell, Schilpzand, & Leavitt, 2010). Ensuring transparency and regular audits of AI-driven evaluations can further enhance their fairness and reliability, building confidence among employees and stakeholders in the system's integrity (Castilla, 2008).

In conclusion, our study demonstrates that LLMs, specifically GPT-4, can serve as reliable and superior alternatives to human raters in performance evaluations, offering higher consistency and reliability. Key findings include strong correlations between GPT and aggregated human ratings and the ability of LLMs to extract meaningful constructs from text-based data, enhancing theoretical understanding in management. These advancements enable management scholars to explore new dimensions of organizational behavior with greater empirical rigor.

Zhao, W. X., Zhou, K., Li, J., Tang, T., Wang, X., et al. 2023. *A Survey of Large Language Models*. http://arxiv.org/abs/2303.18223.

Zhou, J., & George, J. M. 2001. When job dissatisfaction leads to creativity: Encouraging the expression of voice. *Academy of Management Journal*, 44(4): 682–696.

Zhou, J., Wang, X. M., Song, L. J., & Wu, J. 2017. Is it new? Personal and contextual influences on perceptions of novelty and creativity. *Journal of Applied Psychology*, 102(2): 180–202.

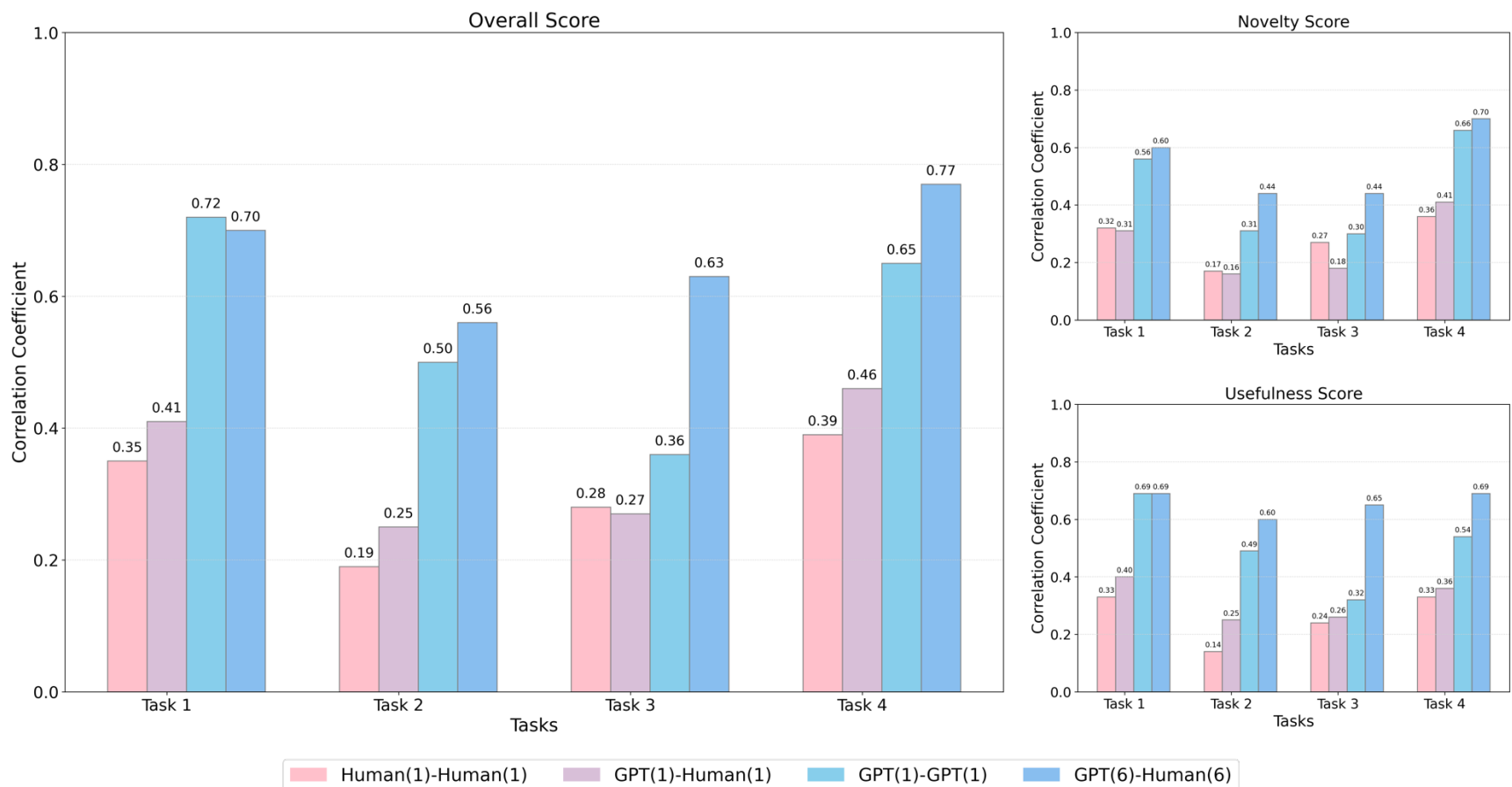

Figure 1. Correlations of Single GPT Ratings and Single Human Ratings

Notes: In these notations, "Human" refers to human ratings and "GPT" refers to GPT-generated ratings. The number in parentheses (e.g., 1) indicates the number of raters combined. For example, Human(1) represents a single human rater's evaluation, while GPT(1) represents a single GPT-generated rating. Additionally, Human-Human indicates the correlation between human ratings, while Human-GPT indicates the correlation between human ratings and GPT ratings.

**Figure 2. Correlations of Single GPT Ratings and Single Human Ratings with Aggregated Ratings**

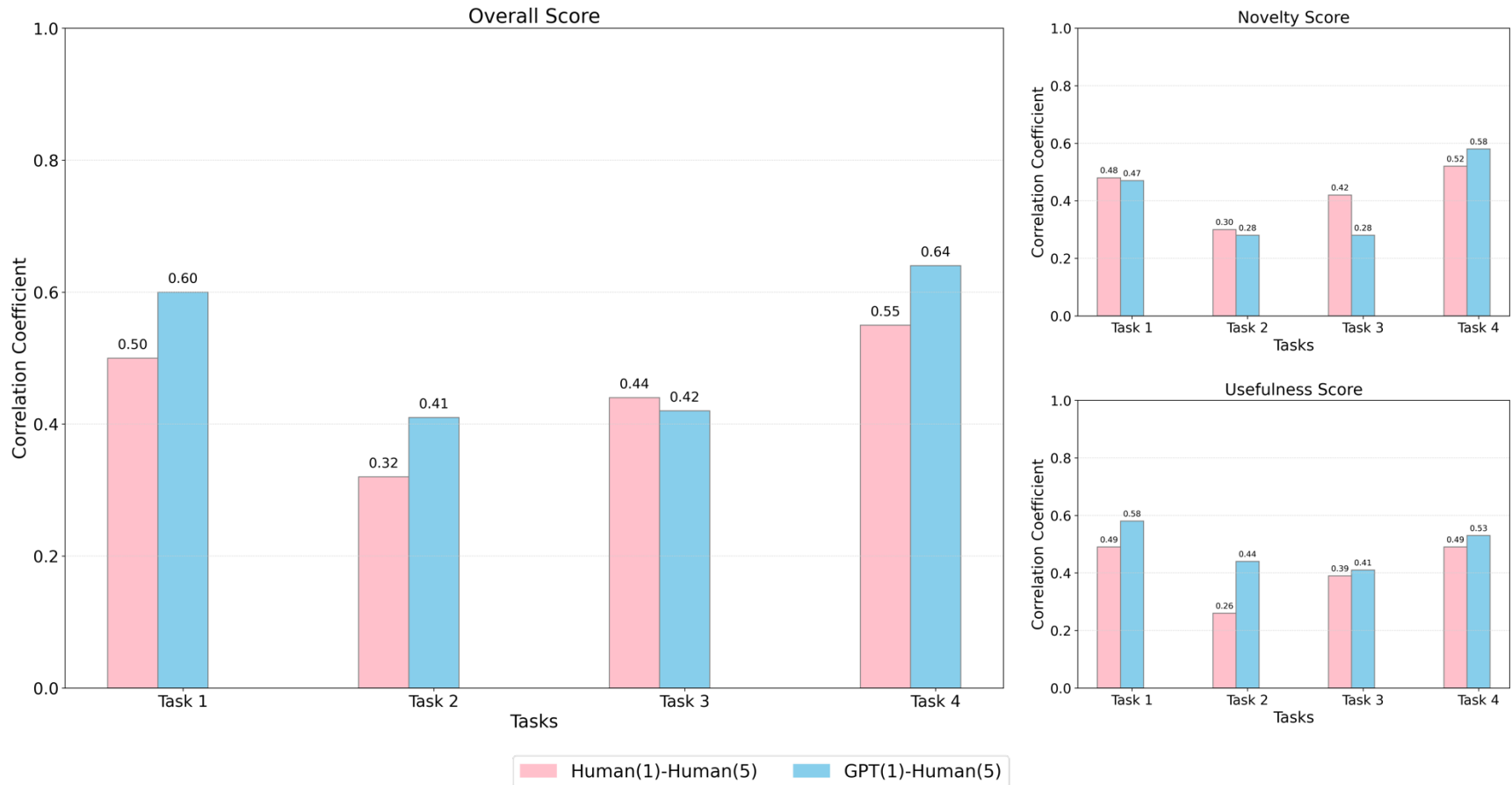

Notes: "Human" refers to human ratings and "GPT" refers to GPT-generated ratings. The number in parentheses (e.g., 1) indicates the number of raters combined. For example, Human(5) represents five human raters' evaluation, while GPT(5) represents five GPT-generated ratings. Additionally, Human-Human indicates the correlation between human ratings, while Human-GPT indicates the correlation between human ratings and GPT ratings.

**Figure 3. Idiosyncratic Variance of Multiple GPT Rating and Multiple Human Rating**

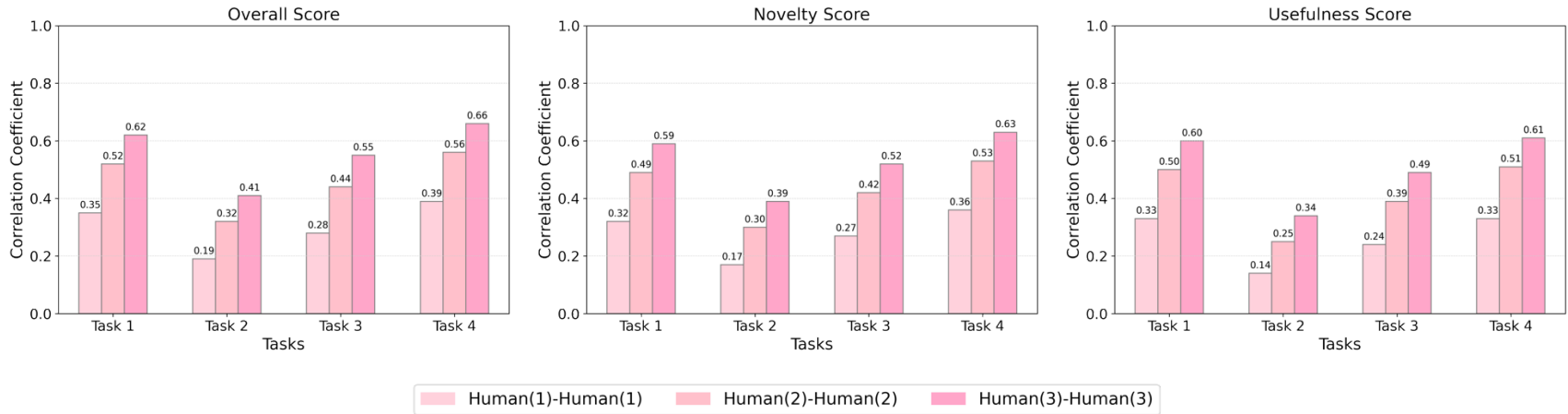

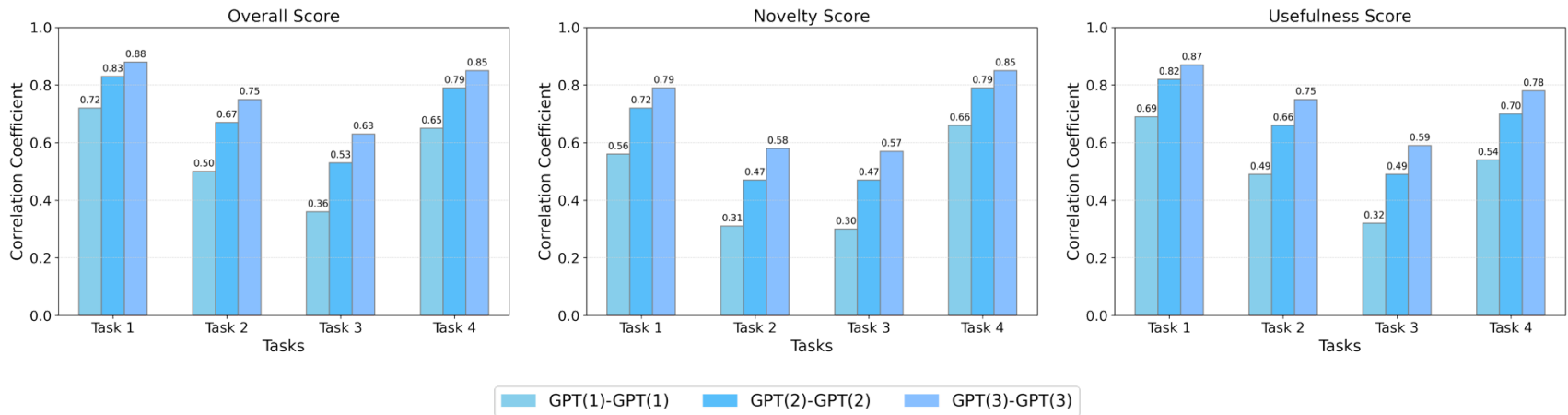

# Figure 4. Accuracy of Multiple GPT Ratings and Multiple Human Ratings

Task 1.

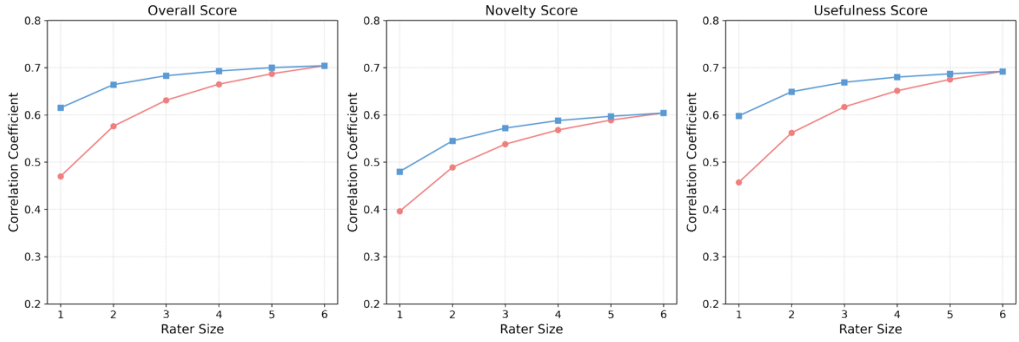

Task 2.

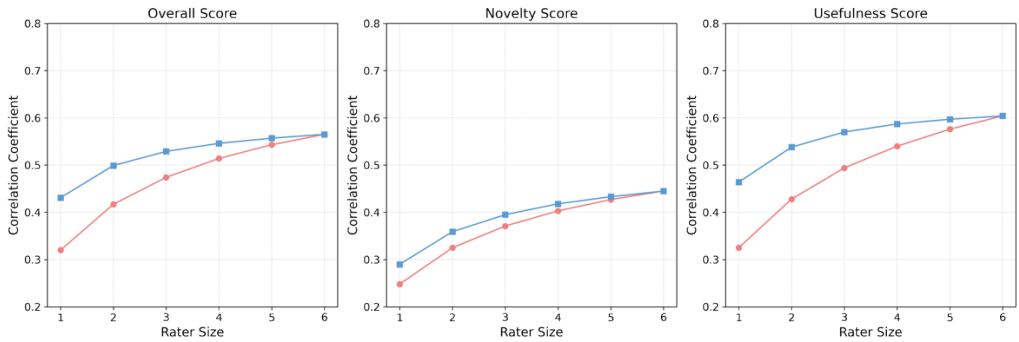

Task 3.

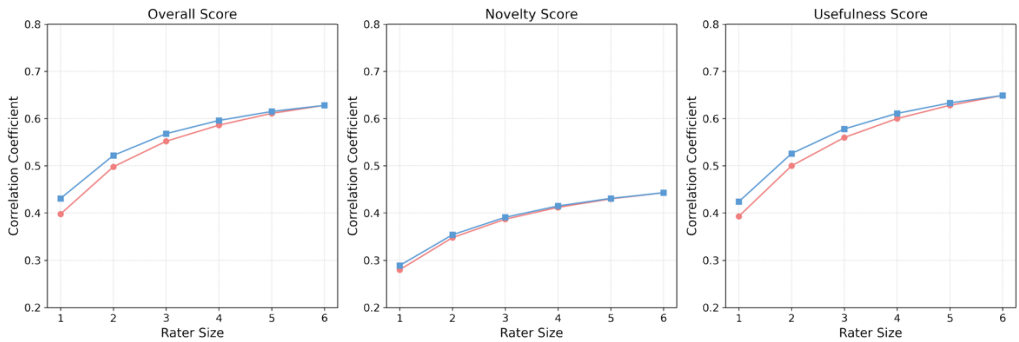

Task 4.

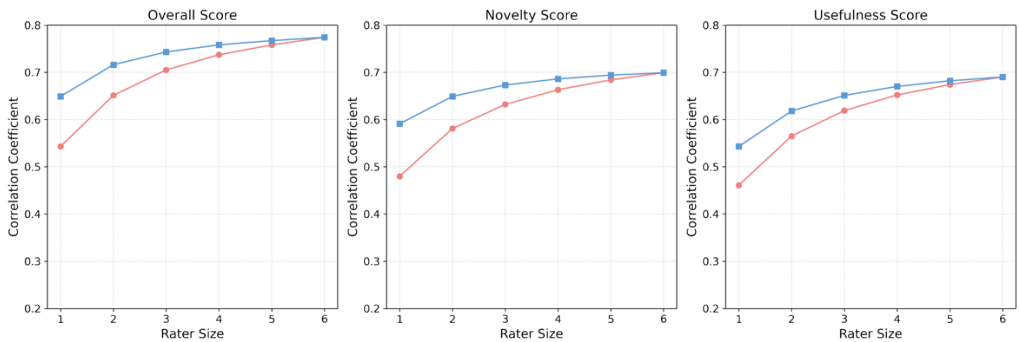

● Human(Rater Size)-GPT(6)　　■ GPT(Rater Size)-Human(6)

**Figure 5. Correlation within Each Rater across Different Dimensions**

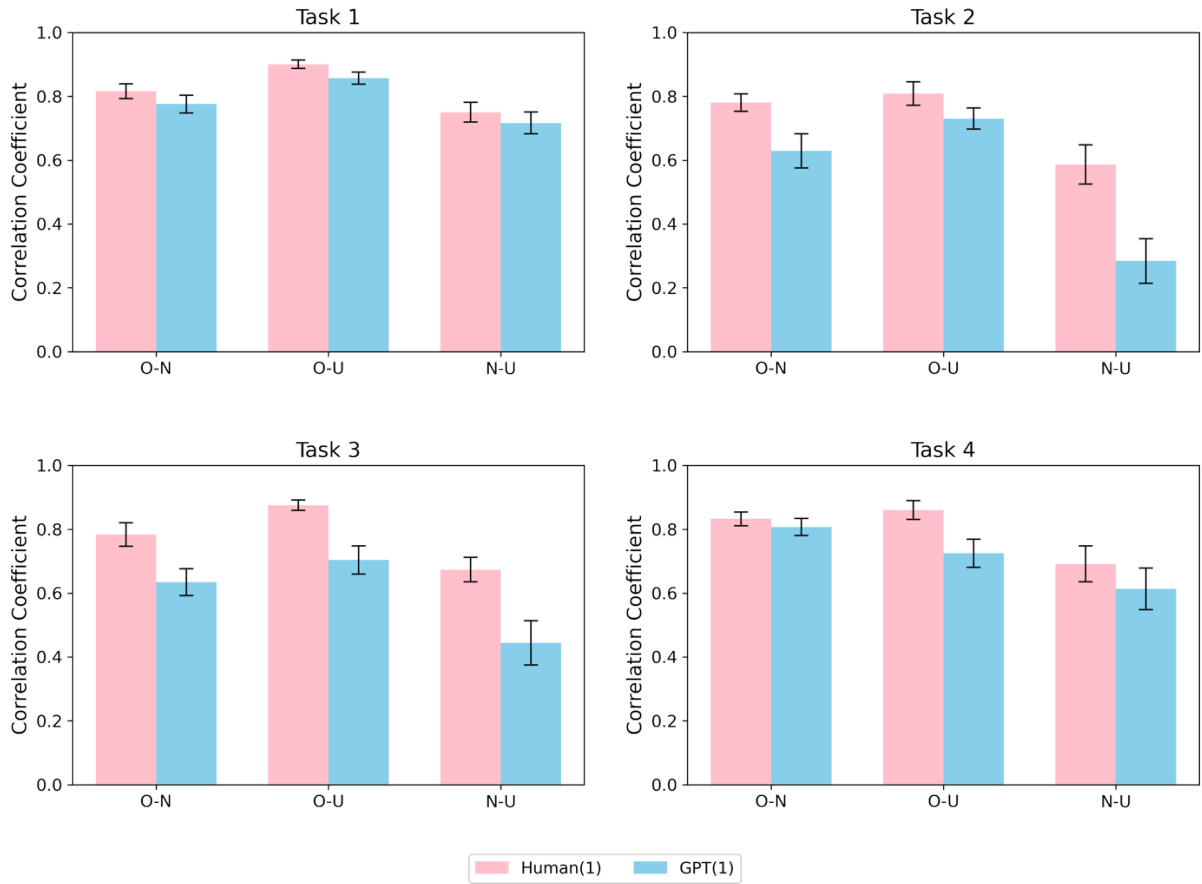

Notes: O = Overall rating dimension; N = Novelty rating dimension; U = Usefulness rating dimension

**Figure 6. Accuracy Comparison of Multiple GPT Ratings at Different Temperatures**

Task 1.
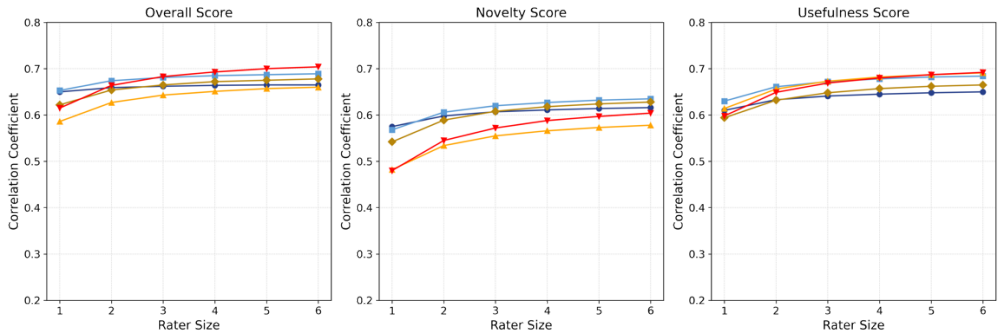

Task 2.
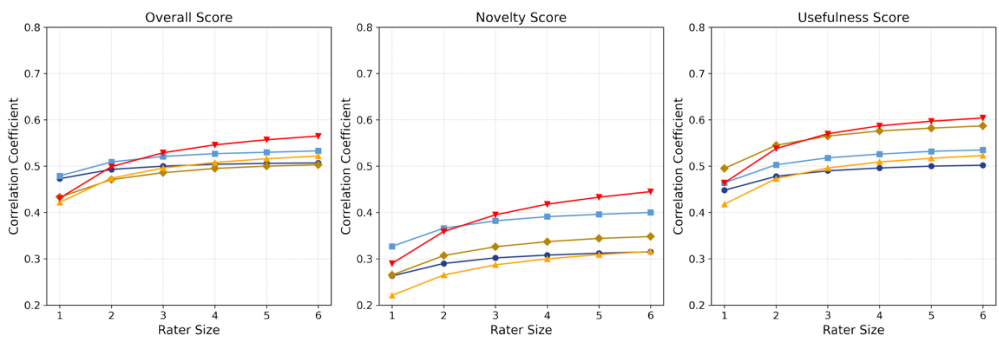

Task 3.
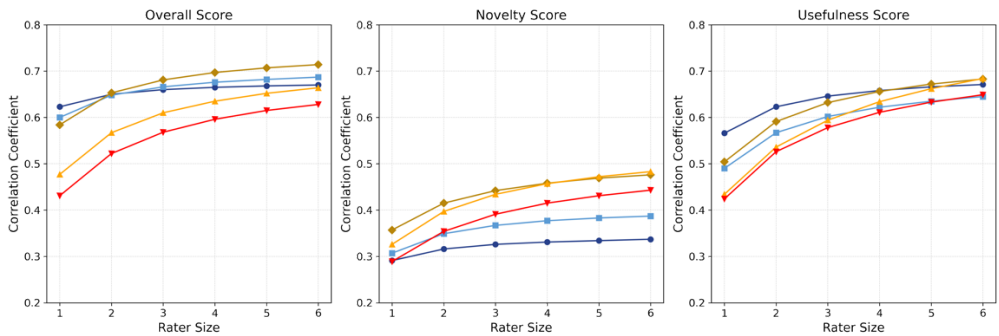

Task 4.
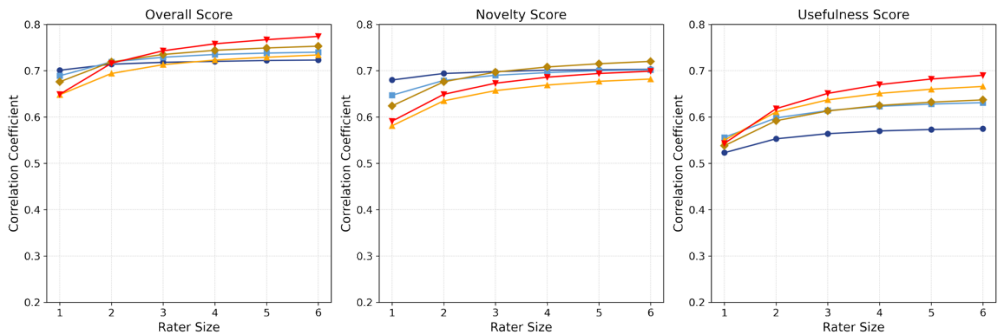

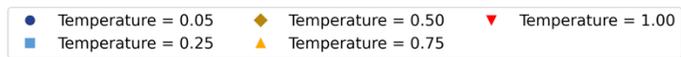

**Figure 7. Research Design of Study 2**

| | Halo Group | Halo Mitigation Group | Control Group |
|---|---|---|---|
| Human Raters | Information provided:<br>**Halo text (Positive/Neutral/Negative) + Performance output.**<br><br>Halo text examples:<br>(Positive) "The growth rate of Employee xxx in recent years has been remarkable, rapidly rising to become a key member of the product team. He earned an MBA from Shanghai Jiao Tong University and has accumulated extensive knowledge and practical experience in the ride-hailing industry, which has enabled him to perform exceptionally well in his role. Under his active participation, the product team has achieved significant performance growth for four consecutive years, far exceeding the company's set targets. Due to his outstanding contributions, Employee xxx has been repeatedly awarded the company's Excellent Employee Award."<br><br>(Neutral) "Employee xxx currently holds the position of Product Specialist at the company. He graduated from a mid-tier university with a major in Marketing and has previous experience as a Product Specialist in a similar company. Since joining the company, he has consistently fulfilled the essential tasks required for his position with a diligent work ethic and steady performance. While he has proven to be reliable in executing daily responsibilities, his performance has yet to demonstrate any particularly noteworthy achievements."<br><br>(Negative) "Employee xxx is an Operations Specialist at the company. He has a high school diploma and has frequently changed jobs since graduation, with most of his work experiences being short-term and at lower levels. Since joining the company, he has often failed to meet performance targets, frequently makes mistakes in daily tasks, and has shown a lukewarm response to suggestions for improving work efficiency and quality. Additionally, he exhibits significant deficiencies in teamwork and communication skills." | Information provided:<br>**Halo text (Positive/Neutral/Negative) + Mitigation instruction + Performance output.**<br><br>Mitigation instruction example:<br>"Please do not let the employees' background information influence you. Evaluate their responses objectively and impartially." | Information provided:<br>**Performance output only.** |
| GPT Raters | Information provided:<br>**Halo text (Positive/Neutral/Negative) + Performance output.** | Information provided:<br>**Halo text (Positive/Neutral/Negative) + Mitigation instruction + Performance output.** | Information provided:<br>**Performance output only.** |

**Figure 8. Halo Effect: Correlation of Scores in Different Groups**

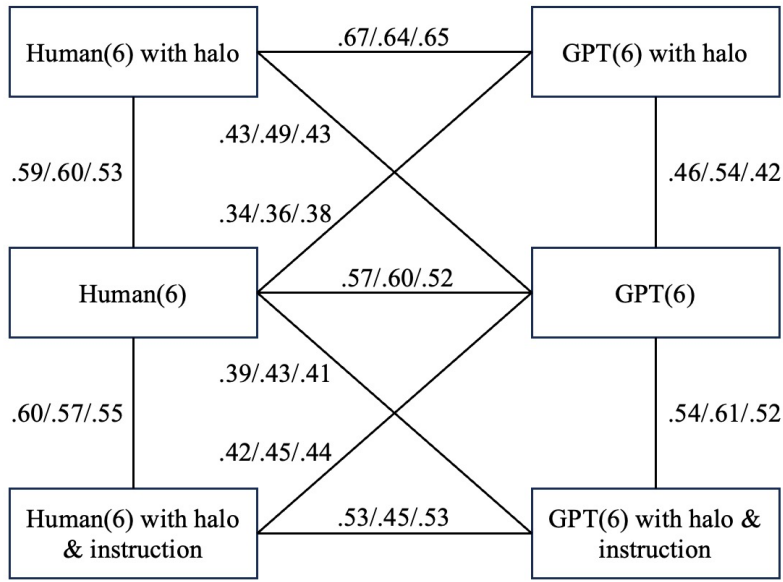

Note: The three numbers on the line sequentially represent the correlations of the three dimensions (i.e., *Overall*, *Novelty*, and *Usefulness*) between the ratings of the two groups at either end of the line

**Table 1. Meta-Analysis: Consistency of Single GPT Ratings and Single Human Ratings**

|  |  |  | 10%CV | 90%CV | 95%LCI | 95%UCI | K |
|---|---|---|---|---|---|---|---|
| Overall | Task 1 | Human(1)-Human(1) | .26 | .44 | .29 | .40 | 15 |
|  |  | GPT(1)-GPT(1) | .72 | .72 | .70 | .74 | 15 |
|  |  | GPT(1)-Human(1) | .39 | .43 | .39 | .44 | 36 |
|  | Task 2 | Human(1)-Human(1) | .09 | .28 | .13 | .24 | 15 |
|  |  | GPT(1)-GPT(1) | .50 | .50 | .47 | .53 | 15 |
|  |  | GPT(1)-Human(1) | .20 | .29 | .22 | .28 | 36 |
|  | Task 3 | Human(1)-Human(1) | .28 | .28 | .24 | .32 | 15 |
|  |  | GPT(1)-GPT(1) | .30 | .42 | .31 | .40 | 15 |
|  |  | GPT(1)-Human(1) | .27 | .27 | .25 | .30 | 36 |
|  | Task 4 | Human(1)-Human(1) | .30 | .48 | .34 | .44 | 15 |
|  |  | GPT(1)-GPT(1) | .65 | .65 | .62 | .67 | 15 |
|  |  | GPT(1)-Human(1) | .46 | .46 | .43 | .48 | 36 |
| Novelty | Task 1 | Human(1)-Human(1) | .21 | .44 | .26 | .39 | 15 |
|  |  | GPT(1)-GPT(1) | .56 | .56 | .53 | .59 | 15 |
|  |  | GPT(1)-Human(1) | .31 | .32 | .29 | .34 | 36 |
|  | Task 2 | Human(1)-Human(1) | .10 | .25 | .12 | .22 | 15 |
|  |  | GPT(1)-GPT(1) | .31 | .31 | .27 | .35 | 15 |
|  |  | GPT(1)-Human(1) | .16 | .16 | .13 | .19 | 36 |
|  | Task 3 | Human(1)-Human(1) | .26 | .28 | .23 | .31 | 15 |
|  |  | GPT(1)-GPT(1) | .30 | .30 | .26 | .34 | 15 |
|  |  | GPT(1)-Human(1) | .14 | .23 | .15 | .21 | 36 |
|  | Task 4 | Human(1)-Human(1) | .25 | .47 | .30 | .42 | 15 |
|  |  | GPT(1)-GPT(1) | .66 | .66 | .63 | .68 | 15 |
|  |  | GPT(1)-Human(1) | .41 | .41 | .38 | .43 | 36 |
| Usefulness | Task 1 | Human(1)-Human(1) | .29 | .38 | .29 | .37 | 15 |
|  |  | GPT(1)-GPT(1) | .69 | .69 | .67 | .71 | 15 |
|  |  | GPT(1)-Human(1) | .33 | .46 | .37 | .42 | 36 |
|  | Task 2 | Human(1)-Human(1) | .10 | .19 | .10 | .19 | 15 |
|  |  | GPT(1)-GPT(1) | .43 | .55 | .45 | .53 | 15 |
|  |  | GPT(1)-Human(1) | .25 | .25 | .22 | .28 | 36 |
|  | Task 3 | Human(1)-Human(1) | .24 | .24 | .20 | .28 | 15 |
|  |  | GPT(1)-GPT(1) | .27 | .36 | .27 | .36 | 15 |
|  |  | GPT(1)-Human(1) | .18 | .33 | .22 | .29 | 36 |
|  | Task 4 | Human(1)-Human(1) | .25 | .42 | .28 | .38 | 15 |
|  |  | GPT(1)-GPT(1) | .54 | .54 | .51 | .57 | 15 |
|  |  | GPT(1)-Human(1) | .32 | .41 | .34 | .39 | 36 |

Notes: CV = credibility interval, LCI = Lower bound of the confidence interval; UCI = Upper bound of the confidence interval; K: Number of correlations included in the meta-analysis.

**Table 2. Meta-Analysis: Accuracy of Single GPT Ratings and Single Human Ratings**

|  |  |  | 10%CV | 90%CV | 95%LCI (Whitener) | 95%UCI (Whitener) | K |
|---|---|---|---|---|---|---|---|
| Overall | Task 1 | Human(1)-Human(5) | .45 | .56 | .44 | .57 | 6 |
|  |  | GPT(1)-Human(5) | .60 | .60 | .58 | .62 | 36 |
|  |  | Human(1)-GPT(5) | .44 | .50 | .44 | .49 | 36 |
|  | Task 2 | Human(1)-Human(5) | .32 | .32 | .25 | .38 | 6 |
|  |  | GPT(1)-Human(5) | .41 | .41 | .39 | .44 | 36 |
|  |  | Human(1)-GPT(5) | .28 | .35 | .29 | .34 | 36 |
|  | Task 3 | Human(1)-Human(5) | .44 | .44 | .38 | .49 | 6 |
|  |  | GPT(1)-Human(5) | .37 | .47 | .39 | .45 | 36 |
|  |  | Human(1)-GPT(5) | .39 | .39 | .36 | .41 | 36 |
|  | Task 4 | Human(1)-Human(5) | .50 | .59 | .49 | .60 | 6 |
|  |  | GPT(1)-Human(5) | .64 | .64 | .62 | .65 | 36 |
|  |  | Human(1)-GPT(5) | .54 | .54 | .52 | .56 | 36 |
| Novelty | Task 1 | Human(1)-Human(5) | .42 | .54 | .41 | .55 | 6 |
|  |  | GPT(1)-Human(5) | .47 | .47 | .45 | .49 | 36 |
|  |  | Human(1)-GPT(5) | .39 | .39 | .37 | .42 | 36 |
|  | Task 2 | Human(1)-Human(5) | .27 | .33 | .23 | .36 | 6 |
|  |  | GPT(1)-Human(5) | .28 | .28 | .25 | .30 | 36 |
|  |  | Human(1)-GPT(5) | .24 | .24 | .21 | .27 | 36 |
|  | Task 3 | Human(1)-Human(5) | .42 | .42 | .36 | .47 | 6 |
|  |  | GPT(1)-Human(5) | .22 | .34 | .25 | .31 | 36 |
|  |  | Human(1)-GPT(5) | .22 | .33 | .24 | .30 | 36 |
|  | Task 4 | Human(1)-Human(5) | .52 | .52 | .46 | .57 | 6 |
|  |  | GPT(1)-Human(5) | .58 | .58 | .56 | .60 | 36 |
|  |  | Human(1)-GPT(5) | .48 | .48 | .45 | .50 | 36 |
| Usefulness | Task 1 | Human(1)-Human(5) | .49 | .49 | .43 | .54 | 6 |
|  |  | GPT(1)-Human(5) | .58 | .58 | .56 | .60 | 36 |
|  |  | Human(1)-GPT(5) | .38 | .53 | .42 | .48 | 36 |
|  | Task 2 | Human(1)-Human(5) | .26 | .26 | .19 | .32 | 6 |
|  |  | GPT(1)-Human(5) | .44 | .44 | .42 | .47 | 36 |
|  |  | Human(1)-GPT(5) | .32 | .32 | .30 | .35 | 36 |
|  | Task 3 | Human(1)-Human(5) | .39 | .39 | .33 | .45 | 6 |
|  |  | GPT(1)-Human(5) | .28 | .54 | .37 | .45 | 36 |
|  |  | Human(1)-GPT(5) | .38 | .38 | .36 | .41 | 36 |
|  | Task 4 | Human(1)-Human(5) | .41 | .57 | .42 | .57 | 6 |
|  |  | GPT(1)-Human(5) | .53 | .53 | .51 | .55 | 36 |
|  |  | Human(1)-GPT(5) | .40 | .51 | .43 | .48 | 36 |

Notes: CV = credibility interval, LCI =Lower bound of the confidence interval; UCI = Upper bound of the confidence interval; K: Number of correlations included in the meta-analysis

**Table 3. Meta-Analysis: Consistency of Multiple GPT Ratings and Multiple Human Ratings**

| | | | 10%CV | 90%CV | 95%LCI (Whitener) | 95%UCI (Whitener) | K |
|---|---|---|---|---|---|---|---|
| Overall | Task 1 | Human(1)-Human(1) | .26 | .44 | .29 | .40 | 15 |
| | | Human(2)-Human(2) | .46 | .57 | .49 | .54 | 45 |
| | | Human(3)-Human(3) | .60 | .63 | .58 | .65 | 10 |
| | | GPT(1)-GPT(1) | .72 | .72 | .70 | .74 | 15 |
| | | GPT(2)-GPT(2) | .83 | .83 | .83 | .84 | 45 |
| | | GPT(3)-GPT(3) | .88 | .88 | .87 | .90 | 10 |
| | Task 2 | Human(1)-Human(1) | .09 | .28 | .13 | .24 | 15 |
| | | Human(2)-Human(2) | .27 | .37 | .29 | .34 | 45 |
| | | Human(3)-Human(3) | .41 | .41 | .36 | .46 | 10 |
| | | GPT(1)-GPT(1) | .50 | .50 | .47 | .53 | 15 |
| | | GPT(2)-GPT(2) | .67 | .67 | .65 | .68 | 45 |
| | | GPT(3)-GPT(3) | .75 | .75 | .73 | .77 | 10 |
| | Task 3 | Human(1)-Human(1) | .28 | .28 | .24 | .32 | 15 |
| | | Human(2)-Human(2) | .44 | .44 | .42 | .46 | 45 |
| | | Human(3)-Human(3) | .55 | .55 | .51 | .58 | 10 |
| | | GPT(1)-GPT(1) | .30 | .42 | .31 | .40 | 15 |
| | | GPT(2)-GPT(2) | .53 | .53 | .51 | .55 | 45 |
| | | GPT(3)-GPT(3) | .63 | .63 | .60 | .66 | 10 |
| | Task 4 | Human(1)-Human(1) | .30 | .48 | .34 | .44 | 15 |
| | | Human(2)-Human(2) | .51 | .61 | .54 | .58 | 45 |
| | | Human(3)-Human(3) | .63 | .69 | .63 | .69 | 10 |
| | | GPT(1)-GPT(1) | .65 | .65 | .62 | .67 | 15 |
| | | GPT(2)-GPT(2) | .79 | .79 | .78 | .80 | 45 |
| | | GPT(3)-GPT(3) | .85 | .85 | .83 | .86 | 10 |
| Novelty | Task 1 | Human(1)-Human(1) | .21 | .44 | .26 | .39 | 15 |
| | | Human(2)-Human(2) | .41 | .57 | .46 | .52 | 45 |
| | | Human(3)-Human(3) | .54 | .65 | .55 | .64 | 10 |
| | | GPT(1)-GPT(1) | .56 | .56 | .53 | .59 | 15 |
| | | GPT(2)-GPT(2) | .72 | .72 | .71 | .73 | 45 |
| | | GPT(3)-GPT(3) | .79 | .79 | .77 | .81 | 10 |
| | Task 2 | Human(1)-Human(1) | .10 | .25 | .12 | .22 | 15 |
| | | Human(2)-Human(2) | .27 | .33 | .27 | .32 | 45 |
| | | Human(3)-Human(3) | .39 | .39 | .34 | .43 | 10 |
| | | GPT(1)-GPT(1) | .31 | .31 | .27 | .35 | 15 |
| | | GPT(2)-GPT(2) | .47 | .47 | .45 | .49 | 45 |
| | | GPT(3)-GPT(3) | .58 | .58 | .54 | .61 | 10 |
| | Task 3 | Human(1)-Human(1) | .26 | .28 | .23 | .31 | 15 |
| | | Human(2)-Human(2) | .42 | .42 | .40 | .44 | 45 |
| | | Human(3)-Human(3) | .52 | .52 | .48 | .56 | 10 |
| | | GPT(1)-GPT(1) | .30 | .30 | .26 | .34 | 15 |
| | | GPT(2)-GPT(2) | .47 | .47 | .45 | .49 | 45 |
| | | GPT(3)-GPT(3) | .57 | .57 | .54 | .61 | 10 |
| | Task 4 | Human(1)-Human(1) | .25 | .47 | .30 | .42 | 15 |
| | | Human(2)-Human(2) | .45 | .61 | .51 | .56 | 45 |
| | | Human(3)-Human(3) | .56 | .70 | .58 | .68 | 10 |
| | | GPT(1)-GPT(1) | .66 | .66 | .63 | .68 | 15 |
| | | GPT(2)-GPT(2) | .79 | .79 | .78 | .80 | 45 |
| | | GPT(3)-GPT(3) | .85 | .85 | .84 | .87 | 10 |
| Usefulness | Task 1 | Human(1)-Human(1) | .29 | .38 | .29 | .37 | 15 |
| | | Human(2)-Human(2) | .50 | .50 | .48 | .52 | 45 |
| | | Human(3)-Human(3) | .60 | .60 | .56 | .63 | 10 |
| | | GPT(1)-GPT(1) | .69 | .69 | .67 | .71 | 15 |
| | | GPT(2)-GPT(2) | .82 | .82 | .81 | .83 | 45 |
| | | GPT(3)-GPT(3) | .87 | .87 | .86 | .88 | 10 |
| | Task 2 | Human(1)-Human(1) | .10 | .19 | .10 | .19 | 15 |
| | | Human(2)-Human(2) | .25 | .25 | .23 | .28 | 45 |
| | | Human(3)-Human(3) | .34 | .34 | .29 | .39 | 10 |
| | | GPT(1)-GPT(1) | .43 | .55 | .45 | .53 | 15 |
| | | GPT(2)-GPT(2) | .66 | .66 | .65 | .68 | 45 |
| | | GPT(3)-GPT(3) | .75 | .75 | .72 | .77 | 10 |
| | Task 3 | Human(1)-Human(1) | .24 | .24 | .20 | .28 | 15 |
| | | Human(2)-Human(2) | .39 | .39 | .37 | .41 | 45 |
| | | Human(3)-Human(3) | .49 | .49 | .45 | .53 | 10 |
| | | GPT(1)-GPT(1) | .27 | .36 | .27 | .36 | 15 |
| | | GPT(2)-GPT(2) | .49 | .49 | .47 | .51 | 45 |
| | | GPT(3)-GPT(3) | .59 | .59 | .56 | .63 | 10 |
| | Task 4 | Human(1)-Human(1) | .25 | .42 | .28 | .38 | 15 |
| | | Human(2)-Human(2) | .48 | .53 | .49 | .53 | 45 |
| | | Human(3)-Human(3) | .61 | .61 | .58 | .64 | 10 |
| | | GPT(1)-GPT(1) | .54 | .54 | .51 | .57 | 15 |
| | | GPT(2)-GPT(2) | .70 | .70 | .69 | .72 | 45 |
| | | GPT(3)-GPT(3) | .78 | .78 | .76 | .80 | 10 |

Notes: CV = credibility interval, LCI = Lower bound of the confidence interval; UCI = Upper bound of the confidence interval; K: Number of correlations included in the meta-analysis

**Table 4. Regression Results for Rater Size and Accuracy of GPT Ratings**

|  | DV = Overall | | DV = Novelty | | DV = Usefulness | |
|---|---|---|---|---|---|---|
|  | Model (1) | Model (2) | Model (1) | Model (2) | Model (1) | Model (2) |
| *Temperature* | -.024** | -.096*** | .003 | -.061** | .023* | -.039* |
|  | (.008) | (.017) | (.010) | (.022) | (.009) | (.019) |
| *Rater Size* | .015*** | .035*** | .017*** | .039*** | .020*** | .050*** |
|  | (.002) | (.007) | (.002) | (.010) | (.002) | (.008) |
| *Rater Size*Rater Size* |  | -.004*** |  | -.005** |  | -.006*** |
|  |  | (.001) |  | (.001) |  | (.001) |
| *Rater Size*Temperature* |  | .021*** |  | .018** |  | .018*** |
|  |  | (.004) |  | (.006) |  | (.005) |
| *Task Dummy* | Yes | Yes | Yes | Yes | Yes | Yes |
| *Intercept* | .622*** | .618*** | .526*** | .517*** | .573*** | .553*** |
|  | (.009) | (.014) | (.012) | (.019) | (.010) | (.016) |
| *Obs.* | 120 | 120 | 120 | 120 | 120 | 120 |
| *Adj_$R^2$* | .880 | .912 | .932 | .942 | .743 | .805 |

Note: Robust standard errors in parentheses. * $p < 0.05$, ** $p < 0.01$, *** $p < 0.001$.

**Table 5. Paired T-test for Different Dimensions**

|  |  | Paired Groups | Halo Type | Mean | SD | 95% CI | T value |
|---|---|---|---|---|---|---|---|
| Overall | GPT | Halo group – Control group | Positive | .32*** | .47 | [.21, .43] | 5.93 |
|  |  |  | Neutral | -.77*** | .46 | [-.88, -.67] | -14.48 |
|  |  |  | Negative | -1.90*** | .70 | [-2.06, -1.73] | -23.41 |
|  |  | Halo mitigation group – Control group | Positive | .18** | .48 | [.07, .29] | 3.19 |
|  |  |  | Neutral | -.59*** | .53 | [-.71, -.47] | -9.55 |
|  |  |  | Negative | -1.26*** | .62 | [-1.41, -1.12] | -17.37 |
|  | Human | Halo group – Control group | Positive | .52*** | .82 | [.34, .71] | 5.52 |
|  |  |  | Neutral | -.09 | .84 | [-.28, .10] | -.95 |
|  |  |  | Negative | -.86*** | .79 | [-1.04, -.68] | -9.40 |
|  |  | Halo mitigation group – Control group | Positive | .57*** | .72 | [.40, .74] | 6.81 |
|  |  |  | Neutral | .14 | .82 | [-.05, .33] | 1.47 |
|  |  |  | Negative | -.46*** | .89 | [-.67, -.26] | -4.46 |
| Novelty | GPT | Halo group – Control group | Positive | .46*** | .78 | [.28, .64] | 5.08 |
|  |  |  | Neutral | -.91*** | .64 | [-1.06, -.77] | -12.34 |
|  |  |  | Negative | -1.76*** | .65 | [-1.91, -1.61] | -23.17 |
|  |  | Halo mitigation group – Control group | Positive | .21* | .77 | [.03, .38] | 2.29 |
|  |  |  | Neutral | -.80*** | .72 | [-.96, -.63] | -9.53 |
|  |  |  | Negative | -1.29*** | .74 | [-1.46, -1.12] | -15.06 |
|  | Human | Halo group – Control group | Positive | .51*** | 1.08 | [.26, .76] | 4.10 |
|  |  |  | Neutral | -.20 | .91 | [-.41, .01] | -1.94 |
|  |  |  | Negative | -.89*** | .88 | [-1.09, -.68] | -8.66 |
|  |  | Halo mitigation group – Control group | Positive | .36** | 1.08 | [.11, .61] | 2.84 |
|  |  |  | Neutral | -.06 | 1.04 | [-.30, .18] | -.51 |
|  |  |  | Negative | -.32* | 1.10 | [-.57, -.06] | -2.48 |
| Usefulness | GPT | Halo group – Control group | Positive | .43*** | .62 | [.29, .57] | 5.96 |
|  |  |  | Neutral | -.75*** | .69 | [-.91, -.59] | -9.47 |
|  |  |  | Negative | -2.01*** | .81 | [-2.20, -1.83] | -21.45 |
|  |  | Halo mitigation group – Control group | Positive | .32*** | .56 | [.18, .45] | 4.81 |
|  |  |  | Neutral | -.65*** | .64 | [-.80, -.51] | -8.84 |
|  |  |  | Negative | -1.47*** | .79 | [-1.65, -1.29] | -16.09 |
|  | Human | Halo group – Control group | Positive | .38** | 1.01 | [.15, .62] | 3.24 |
|  |  |  | Neutral | -.13 | .85 | [-.33, .06] | -1.35 |
|  |  |  | Negative | -.85*** | .91 | [-1.06, -.64] | -7.99 |
|  |  | Halo mitigation group – Control group | Positive | .41*** | .87 | [.21, .62] | 4.09 |
|  |  |  | Neutral | .07 | .86 | [-.12, .27] | .74 |
|  |  |  | Negative | -.46*** | 1.02 | [-.70, -.22] | -3.88 |